# Extension of Path Probability Method to Approximate Inference over Time

A Thesis

Submitted For the Degree of

**𝕸aster of 𝕾cience (𝕰ngineering)**

in the Computer Science and Engineering

by

**Vinay Jethava**

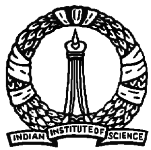

Computer Science and Automation

Indian Institute of Science

BANGALORE – 560 012

May 2009

TO

*My parents and my teachers*

# Abstract

There has been a tremendous growth in publicly available digital video footage over the past decade. This has necessitated the development of new techniques in computer vision geared towards efficient analysis, storage and retrieval of such data. Many mid-level computer vision tasks such as segmentation, object detection, tracking, etc. involve an inference problem based on the video data available. Video data has a high degree of spatial and temporal coherence. The property must be intelligently leveraged in order to obtain better results.

Graphical models, such as Markov Random Fields, have emerged as a powerful tool for such inference problems. They are naturally suited for expressing the spatial dependencies present in video data, It is however, not clear, how to extend the existing techniques for the problem of inference over time. This thesis explores the Path Probability Method, a variational technique in statistical mechanics, in the context of graphical models and approximate inference problems. It extends the method to a general framework for problems involving inference in time, resulting in an algorithm, *DynBP*. We explore the relation of the algorithm with existing techniques, and find the algorithm competitive with existing approaches.

The main contribution of this thesis are the extended GBP algorithm, the extension of Path Probability Methods to the DynBP algorithm and the relationship between them. We have also explored some applications in computer vision involving temporal evolution with promising results.

# Contents

# Notation and Abbreviations

| Symbol | Definition |
| --- | --- |
| $G$ | undirected graph |
| $V$ | vertex or node set of a graph |
| $E$ | edge set of graph |
| $\alpha$ | cluster of a region graph |
| $c_\alpha$ | counting number of cluster $\alpha$ |
| $\mathbf{x}_\alpha$ | state of the cluster |
| $b_\alpha(\mathbf{x}_\alpha)$ | belief that cluster $\alpha$ has state $\mathbf{x}_\alpha$ |

# Chapter 1

# Introduction

There has been an exponential growth of digital video data publicly available during the past decade. This has posed new problems dealing with storage, analysis, classification, retrieval of videos. Therefore, making *interesting* observations about video data has become paramount in the field of computer vision. We have, so far, not clearly specified what we mean by an interesting observation. An interesting observation in video is largely dependent on the context of the problem. For example, in surveillance videos, detecting motion in video or whether a person is present or not might be important, whereas, in a video shot from a mobile phone, one might be interested in cleaning up the jitter that might be present in the video initially before archiving.

Traditional image processing techniques have been successfully applied to video data. However, videos exhibit a strong degree of causality, i.e., the natural progression of the video in time. Several existing methods have successfully exploited the temporal information available in videos.

One of the contributions of this thesis is a framework for inference for probabilistic models which exhibit causality and spatial correlation. This analysis is well-suited for inference in many computer vision problems.

Graphical models [35, 18] and belief propagation algorithms [25, 1, 43] have emerged as powerful tools for a variety of inference problems due to ease of implementation and applicability. They have proven of great appeal in computer vision as they can implicitly capture the spatial correlations in image data. Belief propagation algorithms using MRF models [9, 8, 42] have been used in a number of computer vision problems such as image restoration [11], tracking [12], stereo [40], inpainting [24], shape-matching [7], etc.

Video data exhibits a strong degree of temporal coherence. Further, the change between successive video frames is usually small enough, (except in case of sharp changes, say between scenes) for most common footage. The data also exhibits a sharp degree of spatial coherence. For example, if a pixel is in motion, nearby pixels are more likely to be in motion.

This thesis focusses on approximate inference over systems which are evolving with time. It also explores three computer vision problems that exhibit a high degree of spatio-temporal coherence and are well suited to application of the inference algorithm, DynBP.

- Moving Object Detection

- Video Denoising

- Dropped frame reconstruction

The thesis is organized as follows: Chapter 2 presents the necessary mathematical background of graphical models and variational algorithms, Chapter 3 we extend the path probability methods present in statistical mechanics and present the resultant algorithm, *DynBP*. We analyse the DynBP algorithm and compare it against existing approaches in Chapter 4. We apply our algorithm to computer vision problems and show the experiments and results in 5. Finally, Chapter 6 presents the conclusions and future work.

# Chapter 2

# Graphical Models

Graphical models are a powerful framework for representing and manipulating probability distributions over sets of random variables. They provide a methodology for solving problems involving thousands of random variables that are linked in complex ways, and have found tremendous application in solving statistical problems in many fields such as bioinformatics [38], computer vision [9], communication [10], speech processing [2], etc.

In this we present a brief introduction to the existing literature in the field before focussing on the techniques that are relevant to this thesis. A more comprehensive description can be found in [18, 26].

## 2.1 Basics of graphical models

Graphical models can be either *directed* ( as known as Bayesian networks) or *undirected*. In a graphical model, the edges of the underlying graph represent the probabilistic dependencies between variables. In this thesis, we consider only the undirected graphical models.

Given a graph $G = (V, E)$, a probabilistic graphical model is formed by associating with each node $s \in V$ a random variable $x_s$ taking values in the sample space $\mathcal{X}$. This sample space can either be a continuum (e.g. $\mathcal{X} = \Re$), or the discrete alphabet $\mathcal{X} = \{0, \ldots, m-1\}$. In this latter case, the underlying sample space $\mathcal{X}^N$ is the set of all $N$ vectors $\mathbf{x} = \{x_s | s \in V\}$ over $m$ symbols, so that $\mathcal{X}^N = m^N$.

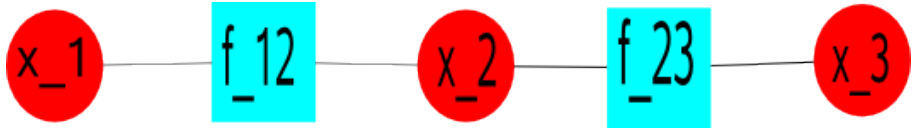

Figure 2.1: A simple factor graph

The joint distribution of an undirected graphical is defined by

$$p(\mathbf{x}) = \frac{1}{Z} \prod_{C \in \mathcal{C}} \psi_C(x_C) \tag{2.1}$$

where $\mathcal{C}$ is the set of maximal cliques in the graph, $\psi_C(x_C)$ is a potential function (a positive, but otherwise arbitrary, real-valued function) on the clique $\mathbf{x}_C$, and $Z$ is the normalization factor

$$Z = \sum_{\mathbf{x}} \prod_{C \in \mathcal{C}} \psi_C(\mathbf{x}_C) \tag{2.2}$$

An undirected graphical model is often represented by a factor graph [25]. A factor graph is a bipartite graph, with its nodes $V \cup \{C \in \mathcal{C}\}$, and an edge between a variable node $x_s \in V$ and a factor node $\psi_C \in \mathcal{C}$ iff the variable is a member of the clique representing the factor, i.e., $x_s \subset \mathbf{x}_C$.

## 2.1.1 Example

Consider the graphical model shown in 2.1.1. The variables nodes $x_1, x_2, x_3$ and the factor nodes are $f_{12}$ and $f_{23}$. The probability distribution is given by

$$p(x_1, x_2, x_3) = \frac{1}{Z} f_{12}(x_1, x_2) f_{23}(x_2, x_3)$$

If each of the variables can take values in $\{0, 1\}$, the partition function is given by

$$Z = \sum_{x_1=0}^{1} \sum_{x_2=0}^{1} \sum_{x_3=0}^{1} f_{12}(x_1, x_2) f_{23}(x_2, x_3)$$

## 2.2 Inference

A problem that arises in many applications of interest is that of estimating the random vector $\mathbf{x} = \{x_s | s \in V\}$ based on a set of noisy observations $\mathbf{y} = \{y_s | s \in V\}$, For instance, in computer vision [11, 9], the vector $\mathbf{x}$ could represent an image defined on a grid, and $\mathbf{y}$ could represent a noisy or blurred version of the image.

We are often interested in the following problems:

1. The *maximum a posteriori* (MAP) estimate corresponding to finding the most likely state $\mathbf{x}$, based on the noisy observations $\mathbf{y}$ - that is:

$$\hat{\mathbf{x}}_{MAP} = \arg \max_{\mathbf{x} \subset \mathcal{X}^N} p(\mathbf{x}|\mathbf{y}) \tag{2.3}$$

   In absence of noisy information, the problem reduces to finding: $\arg \max_{\mathbf{x} \subset \mathcal{X}^N} p(\mathbf{x})$.

2. Computing the marginal distribution of a subset of variables. For the example in 2.1.1, one might be interested in $p(x_1)$, i.e.,

$$p(x_1 = 0) = \sum_{x_2=0}^{1} \sum_{x_3=0}^{1} f_{12}(x_1 = 0, x_2) f_{23}(x_2, x_3)$$

3. Computing the partition function $Z = \sum_{\mathbf{x}} \prod_{C \in \mathcal{C}} \psi_C(\mathbf{x}_C)$

Several methods have been developed solve the problem exactly or approximately.

- Exact inference

- Sampling Methods

- Variational Methods

### 2.2.1 Exact Inference

Exact inference computes the quantity of interest, say the marginal distribution at a node, by appealing to the distributive law [1]. For example, in the factor graph given in

Example 2.1.1, the computation of the marginal at $x_1$ would be according

$$
\begin{aligned}
p(x_1) &= \frac{1}{Z} \sum_{x_2} f_{12}(x_1, x_2) \sum_{x_3} f_{23}(x_2, x_3) \\
&= \frac{1}{Z} \sum_{x_2} f_{12}(x_1, x_2) m_{3 \to 2}(x_2)
\end{aligned}
$$

The order of elimination of the variables is critical to the performance of the algorithm and considerable work has been done in identifying the order.

A common approach is *graph triangulation*, by inserting additional edges, resulting in the junction tree algorithm [16]. A junction tree has the *running intersection* property: If a node appears in any two cliques in the tree, it appears in all cliques that lie on the path between the two cliques. In a junction tree, because of the running intersection property, local consistency implies global consistency.

The main difficulty with the exact approach is that the computation cost in exponential in the size of largest clique in the graph. Thus, exact inference is not possible on large sized graphs. This necessitates alternative methods, which can handle large graph sizes.

**Belief Propagation (BP) and variants**

Belief propagation has emerged as a powerful technique for approximate inference in graphical models. On graphs without cycles, belief propagation is exact and can be viewed as an application of the distributive law with multiple elimination order [35, 1].

It performs exact inference on graphs with cycles. In dense graphs, it provides an approximate solution by passing messages between factor nodes and variable nodes.

The general update equations for the messages for the *sum-product* algorithm [25] are given as

$$
m_{a \to i}(x_i) = \sum_{\vec{x}_a \setminus x_i} f_a(x_a) \prod_{x_j \in \vec{x}_a} m_{j \to a}(x_j) \tag{2.4}
$$

$$
m_{i \to a}(x_i) = \prod_{h \in N(x) \setminus a} m_{h \to i}(x_i) \tag{2.5}
$$

Figure 2.2: The sum-product algorithm update equations (a) message from variable $i$ to factor $a$ based on other neighbouring factors $N(i) - \{a\}$, $m_{i \to a}(x_i) = \prod_{h \in N(x) \backslash a} m_{h \to i}(x_i)$ (b) message from factor $a$ to variable $i$ based on other neighbouring variables $N(a) - \{i\}$, $m_{a \to i}(x_i) = \sum_{\vec{x}_a \backslash x_i} f_a(x_a) \prod_{x_j \in \vec{x}_a} m_{j \to a}(x_j)$

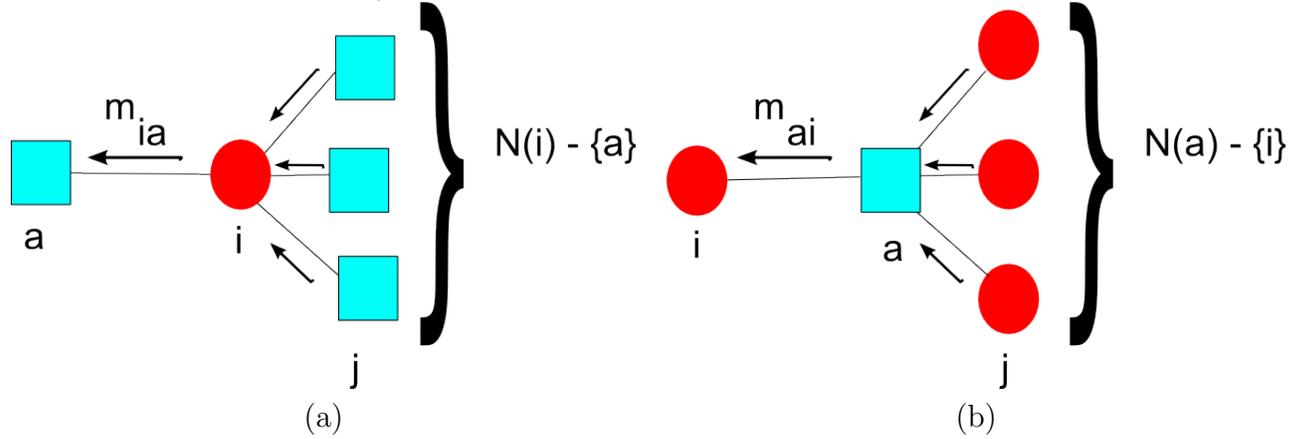

(a)  (b)

$$b(x_i) \quad \propto \quad \prod_{f \in \mathcal{N}(x_i)} m_{f \to i}(x_i) \tag{2.6}$$

where $m_{a \to i}(x_i)$ $(m_{i \to a}(x_i))$ is the message from (to) factor node $a$ to (from) variable node $i$, and $b(x_i)$ is the marginal distribution at variable node $i$.

Wainwright etal. [43] presented an alternative approach which obtains the results for a graph with cycles, as appropriately reweighted convex combination of instances of classic belief propagation runs on trees. Yedidia etal. [49] used *region based free energy approximations* to obtain an algorithm, GBP, with better convergence properties. Weiss etal. [44] have recently investigated the relationship of BP with linear programming.

### 2.2.2 Sampling Methods

Sampling algorithms such as *importance sampling* and *Markov Chain Monte Carlo* methods are an alternative approach for solving approximate inference problems [28, 4]. The algorithms exploit the *Markov blanket* property of graphical models: simply, that conditioning a node on its markov blanket ( which is just the set of its neighbours in the undirected graphical model), renders it independent of all other variables.

The method maintains a proposal distribution from which samples are drawn and accepted or rejected so that the resulting distribution matches the requirements. The *Metropolis Hastings* algorithm draws the samples based on the current state $\mathbf{x}^{(t)}$ s.t. the distribution is $q(\mathbf{x}^{(t+1)}|\mathbf{x}^{(t)})$, and the states $x^{(1)}, x^{(2)}, \ldots$ form a markov chain.

The *Gibbs* algorithm samples a distribution by replacing the value of one of the variables by drawing from a distribution conditioned on the remaining variables. For example, the sample distribution $p(\mathbf{x}) = p(x_1, x_2, \ldots, x_n)$ is sampled at each time to get the new iterate as drawn from $x_i^{(t+1)} \sim p(x_i|\mathbf{x}^{(\mathbf{t})}\backslash i)$ where $\mathbf{x}^{(\mathbf{t})}\backslash i$ is the value of the remaining variables. Similarly, for the next iteration, another index $j$ is chosen based on some order, and the value $x_j^{(t+2)}$ updated based on $\mathbf{x}_{\backslash j}^{(t+1)}$.

The algorithms are simple to implement and provide theoretical guarantees of convergence. However, the convergence is often slow. Newer methods have been developed which ameliorate the problems to an extent: *slice sampling* [30] which adapts the step size in the Metropolis algorithm to match the characteristics of the distribution, Hybrid Monte Carlo algorithm [29] which can make large changes to system state while keeping the rejection probability small.

Sampling methods are most commonly used in case of arbitrary graphs with no structure.

### 2.2.3 Variational Methods

Variational methods refers to techniques which obtain a solution by posing it as the result of an optimization problem. The finite element method [41], maximum entropy estimation [19], mean field methods [31] are instances of variational methods. In this thesis, we focus on the *Naive Mean field Method* in statistical mechanics.

#### Naive Mean Field Method

This method approximates an intractable distribution $p(\mathbf{x}) = p(x_1, \ldots, x_n)$ by a distribution $q(\mathbf{x})$ belonging to a tractable class of distributions $\mathcal{Q}$ which minimizes some distance measure $D(p, q)$ within the class $\mathcal{Q}$. The Kullback Liebler divergence $KL(q||p)$ is often chosen as the distance measure for tractable computation, given by

$$KL(q||p) = \sum_{\mathbf{x}} q(x) \ln \frac{q(\mathbf{x})}{p(\mathbf{x})} \quad (\geq 0) \tag{2.7}$$

The method was originally developed in statistical mechanics dealing with Ising spin systems. The spin systems have a probability distribution given by the Boltzmann distribution

$$P(\mathbf{x}) = \frac{e^{-H(\mathbf{x})}}{Z} \tag{2.8}$$

where $\mathbf{x} = \{x_1, \ldots, x_n\}$ are binary (spin) variables $x_i \in \pm 1$ and

$$H(\mathbf{x}) = -\sum_{i,j} J_{ij} x_i x_j - \sum_i h_i x_i \tag{2.9}$$

The partition function is given $Z = \sum_{\mathbf{x}} e^{-H(\mathbf{x})}$. The direct computation of the partition function involves $O(2^n)$ computations, which is not possible for real large scale systems. The method attempts to compute an approximation to the *free energy* $-\ln Z$ as follows:

$$KL(q||p) = \ln Z + E(q) - S(q) \tag{2.10}$$

where $E(q)$ is the variational energy given by $E(q) = \sum_{\mathbf{x}} q(\mathbf{x}) H(\mathbf{x})$ and $S[q]$ is the entropy of the distribution $q$ given by $S(q) = -\sum_{\mathbf{x}} q(\mathbf{x}) \ln q(\mathbf{x})$.

The mean field approximation considers the distribution $q(\mathbf{x})$ from the class $\mathcal{Q}$ of product distributions, i.e.,

$$q(\mathbf{x}) = \prod_{i=1}^n q_j(x_j) \tag{2.11}$$

For the enthalpy function $H(\mathbf{x})$ in (2.9), the above equation may be rewritten in terms of variational parameters $m_j$ as

$$q_j(x_j; m_j) = \frac{(1 + x_j m_j)}{2} \tag{2.12}$$

where $m_j$ are identified as expectations $m_j = E_q[x_j]$. The variational energy $E(q)$ and the entropy $S(q)$ can be simplified as:

$$S(q) = -\sum_i \left\{ \frac{1+m_i}{2} \ln \frac{1+m_i}{2} + \frac{1-m_i}{2} \ln \frac{1-m_i}{2} \right\} \quad (2.13)$$

$$E(q) = -\sum_{i,j} J_{ij} m_i m_j - \sum_i h_i m_i \quad (2.14)$$

The *variational free energy* F(q), is given by

$$F(q) = E(q) - S(q) \quad (2.15)$$

and it is an upper bound on the true free energy $-\ln Z$ as

$$KL(q||p) \geq 0 \quad \text{(Using (2.10))}$$
$$\Rightarrow -\ln Z \leq E(q) - S(q) = F(q) \quad (2.16)$$

Minimizing the variational free energy $F(q)$ w.r.t. the parameters $m_j$ yields the set of $n$ *mean field equations*, given as:

$$m_i = \tanh\left(\sum_j J_{ij} m_j + h_i\right), \quad \forall i \in 1, \ldots, n \quad (2.17)$$

The mean field method uses the independence of spin probabilities, i.e., $E_q[x_i x_j] = E_q[x_i] E_q[x_j]$, in order to obtain a tractable approximation to the computation of the free energy. We discuss the Bethe-Pierls approximation in the following section, which considers joint pair beliefs in addition to node beliefs.

## Bethe-Pierls Approximation

The Bethe-Pierls approximation derives the variational free energy in terms of single node beliefs $q_i(x_i)$ and pair beliefs $q_{ij}(x_i, x_j)$ with the normalization condition

$$\sum_j q_{ij}(x_i, x_j) = q_i(x_i)$$

The average energy is given by

$$E(q) = -\sum_{ij} J_{ij} q_{ij}(x_i, x_j) - \sum_i h_i q_i(x_i) \tag{2.18}$$

The average energy is exact if the node beliefs $q_{ij}(x_i, x_j)$ and $q_i(x_i)$ are exact [51]. The entropy is given by the approximation

$$S_{Bethe}(q) = -\sum_{x_i x_j} q_{ij}(x_i, x_j) \ln q_{ij}(x_i, x_j) + \sum_i (d_i - 1) \sum_{x_i} q_i(x_i) \ln q_i(x_i) \tag{2.19}$$

where $d_i$ denotes the *degree* of variable node $i$, i.e., the number of factor nodes connected to node $i$. The entropy expression is exact for singly-connected graphs [51], and the overall belief can be expressed as

$$b(\mathbf{x}) = \frac{\prod_{i,j} q_{ij}(x_i, x_j)}{\prod_i q_i(x_i)} \tag{2.20}$$

The variational free energy is given by $G_{Bethe} = E(q) - S_{Bethe}(q)$. It can be shown that for a singly-connected graph, the beliefs obtained by Belief Propagation correspond to global minima of the bethe free energy. Further, for any general graph, the set of beliefs give a BP fixed point if and only if they are local stationary points of the Bethe free energy [51].

Other methods such as the Cluster Variation Method (CVM) [20, 27] consider hierarchy of localized clusters and approximate the system entropy in terms of the distributions on the localized clusters.

In the following section, we present a brief description of the Cluster Variation Method. An extended review of the method and its applications is found in [36].

## 2.3 Cluster Variation Method

The cluster variation method is a generalization of the Bethe-Pierls approximation. It computes the approximate free energy in a statistical system under equilibrium conditions.

The variational free energy is given by

$$F = -\ln Z = \min_q \mathcal{F}(q) = \min_q \sum_{\mathbf{x}} q(\mathbf{x})H(\mathbf{x}) + q(\mathbf{x})\ln q(\mathbf{x}) \qquad (2.21)$$

The underlying idea is to treat the energy term in (2.21) exactly and to approximate by means of a truncated cumulant expansion. A *cluster* or *region* $\mathbf{x}_\alpha \subset \mathbf{x}$ is a subset of a factor graph, such that, is a factor node $f_a$ belongs to $\mathbf{x}_\alpha$, then all variable nodes $x_i$ connected to $f_a$ also belong to $\mathbf{x}_\alpha$. Given a cluster $\mathbf{x}_\alpha$, the energy $H_\alpha(\mathbf{x}_\alpha)$ and the probability distribution on the cluster, $q_\alpha(\mathbf{x}_\alpha)$, are defined as:

$$H_\alpha(\mathbf{x}_\alpha) = \sum_{a \in \alpha} H_a(\mathbf{x}_a) \qquad (2.22)$$

$$q_\alpha(\mathbf{x}_\alpha) = \sum_{\mathbf{x} \setminus \mathbf{x}_\alpha} q(\mathbf{x}) \qquad (2.23)$$

where $H_a(\mathbf{x}_a)$ denotes the energy contribution of factor node $f_a$. The cluster entropy, $S_\alpha$, is given by

$$S_\alpha = -\sum_{\mathbf{x}_\alpha} q_\alpha(\mathbf{x}_\alpha)\ln q_\alpha(\mathbf{x}_\alpha) \qquad (2.24)$$

The entropy cumulants have the following relation any cluster $\alpha$ and all its sub-clusters $\beta \subseteq \alpha$,

$$S_\alpha = \sum_{\beta \subset \alpha} \tilde{S}_\beta \qquad (2.25)$$

where the entropy cumulant $\tilde{S}_\beta$ is given by means of a Möbius inversion as

$$\tilde{S}_\beta = \sum_{\alpha \subseteq \beta} (-1)^{n_\alpha - n_\beta} S_\alpha \qquad (2.26)$$

where $n_\alpha$ denotes the number of variable nodes in cluster $\alpha$. The variational free energy in (2.21) can then be written as

$$\mathcal{F}(q) = \sum_{\mathbf{x}} q(\mathbf{x}) H(\mathbf{x}) - \sum_{\beta} \tilde{S}_{\beta} \qquad (2.27)$$

where the second summation is over all possible clusters.

The above equation is exact. An approximation is made by choosing a set, $\mathcal{R}$, of maximal clusters and their sub-clusters such that each factor node is present in at least one cluster. Then, the cumulant is truncated by retaining only the terms of the cumulant present in $\mathcal{R}$.

$$\sum_{\beta} \tilde{S}_{\beta} \simeq \sum_{\beta \in \mathcal{R}} \tilde{S}_{\beta} = \sum_{\alpha \in \mathcal{R}} c_{\alpha} S_{\alpha} \qquad (2.28)$$

,

where the coefficients $c_{\alpha}$ are known as Möbius numbers and satisfy

$$\sum_{\beta \subseteq \alpha \in \mathcal{R}} c_{\alpha} = 1 \quad \forall \alpha \in \mathcal{R} \qquad (2.29)$$

The free energy in (2.21) is approximated as

$$\mathcal{F}(q_{\alpha}, \, \alpha \in \mathcal{R}) = \sum_{\alpha \in \mathcal{R}} c_{\alpha} \mathcal{F}_{\alpha}(q_{\alpha}) \qquad (2.30)$$

where $\mathcal{F}_{\alpha}(q_{\alpha})$ is the cluster free energy for $\alpha$, given by

$$\mathcal{F}_{\alpha}(q_{\alpha}) = \sum_{\mathbf{x}_{\alpha}} q_{\alpha}(\mathbf{x}_{\alpha}) H_{\alpha}(\mathbf{x}_{\alpha}) + q_{\alpha}(\mathbf{x}_{\alpha}) \ln q_{\alpha}(\mathbf{x}_{\alpha}) \qquad (2.31)$$

The method involves minimization of the approximate free energy in (2.30) w.r.t. the cluster probability distributions $\{q_{\alpha}, \, \alpha \in \mathcal{R}\}$, subject to the constraints,

$$\sum_{\mathbf{x}_{\alpha}} q_{\alpha}(\mathbf{x}_{\alpha}) \;\; = \;\; 1 \; \forall \alpha \in \mathcal{R} \quad \text{(normalization)} \qquad (2.32)$$

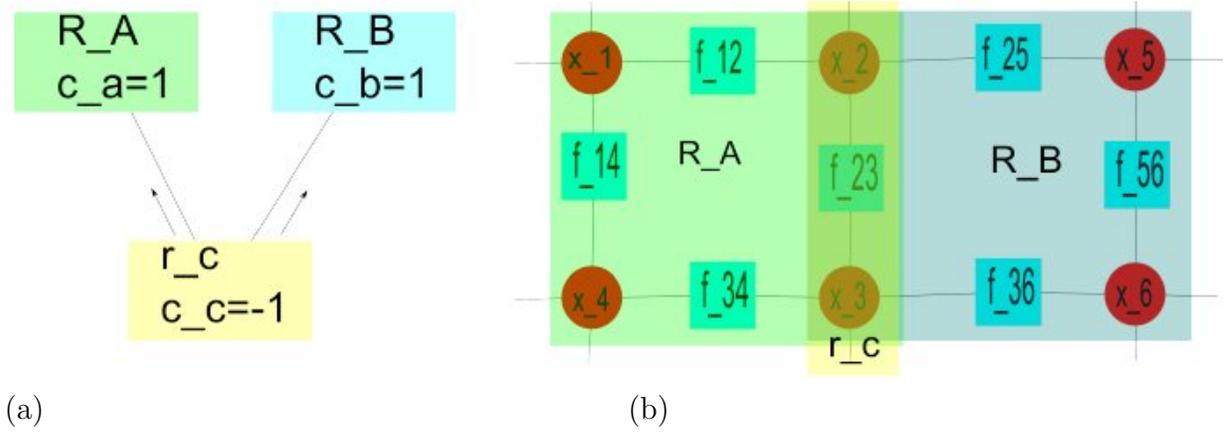

(a)                                                    (b)

Figure 2.3: (b) A factor graph with a set of clusters with associated counting numbers. The maximal clusters are $A$ and $B$, with sub-cluster $c$, with the relation, $c \subset A$ and $c \subset B$. (a) A valid region graph with the associated counting numbers.

$$\sum_{\mathbf{x}_\alpha \setminus \mathbf{x}_\beta} q_\alpha(\mathbf{x}_\alpha) \;=\; q_\beta(\mathbf{x}_\beta) \; \forall \mathbf{x}_\beta, \; \beta \subset \alpha \in \mathcal{R} \quad \text{(compatibility)} \qquad (2.33)$$

### 2.3.1 Example

We consider the factor graph shown in Figure 2.3.1. The probability distributions on the maximal clusters $A$ and $B$ and the common sub-cluster $c$, should satisfy the normalization and compatibility constraints. For example, the normalization constraint for cluster $A$ is given by

$$\sum_{x_1, x_2} q_A(x_1, x_2) = 1$$

and the compatibility constraint between cluster $A$ and sub-cluster $c$ is given by

$$\sum_{x_1} q_A(x_1, x_2) = q_c(x_2) \; \forall x_2$$

## 2.4 Generalized Belief Propagation (GBP)

Yedidia etal. [48, 51] investigated the relationship between the Belief Propagation algorithm and the variational methods. It has been shown that the belief propagation attempts to solve the fixed points of the variational free energy in (2.21) using the Bethe-Pierls approximation, where the big regions consist of individual factor nodes with their associated variable nodes; and the small regions consist of individual variable nodes, thus, establishing a close relationship between the two approaches.

Yedidia etal. [49, 50] developed a new iterative update algorithm, Generalized Belief Propagation (GBP), closely related to the Cluster Variation Method. They proposed the idea of *region-based free energy approximations* which simplified the constraints in (2.32) to counting constraints only on the factor and variable nodes, i.e.,

$$\sum_{\alpha, a \in \alpha} c_\alpha = 1 \ \forall a \quad \text{(factor node)} \tag{2.34}$$

$$\sum_{\alpha, s \in \alpha} c_\alpha = 1 \ \forall s \quad \text{(variable node)} \tag{2.35}$$

Once, an appropriate region graph is constructed which satisfies (2.34), the region beliefs are updated according to the contribution from messages from the neighbours until the beliefs converge. We now describe the parent-to-child version of the GBP algorithm.

### 2.4.1 Parent-to-child algorithm

The *parent-to-child* algorithm passes messages from the parent region to the child region. Each region $R$ has a belief $b_R(\mathbf{x}_R)$ given by

$$b_R(\mathbf{x}_R) \propto \prod_{a \in R} f_a(\mathbf{x}_a) \cdot \left( \prod_{P \in \mathcal{P}(R)} m_{P \to R}(\mathbf{x}_R) \right) \cdot \left( \prod_{D \in \mathcal{D}(R)} \prod_{P' \in \mathcal{P}(D) \setminus \mathcal{E}(R)} m_{P' \to D}(\mathbf{x}_D) \right) \tag{2.36}$$

where $\mathcal{P}(R)$ is the set of regions that are parents to region $R$, $\mathcal{D}(R)$ is the set of all regions that are descendants of region $R$, $\mathcal{E}(R) = R \cup \mathcal{D}(R)$ is the set of all regions that are descendants of $R$ and also region $R$ itself, and $\mathcal{P}(D) \setminus \mathcal{E}(R)$ is the set of all regions that are parents of region $D$ except for region $R$ itself or those those regions that are also descendants of region $R$.

The message update is given by

Figure 2.4: An example of a region graph. The belief at region $R$ is given by $b_R \propto m_{A \to R} m_{B \to R} m_{C \to E} m_{C \to H} m_{F \to H} \prod_{a \in R} f_a(\mathbf{x}_a)$. The parent-to-child message update from $R$ to $E$ is given by $m_{R \to E}(\mathbf{x}_E) := \frac{\sum_{\mathbf{x}_R \setminus \mathbf{x}_E} m_{A \to R}(\mathbf{x}_R) m_{B \to R}(\mathbf{x}_R) \prod_{a \in R \setminus E} f_a(\mathbf{x}_a)}{m_{D \to G}(\mathbf{x}_G)}$.

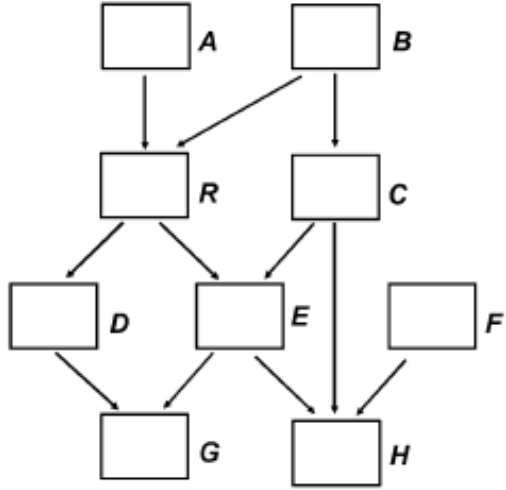

$$m_{P \to R}(\mathbf{x}_R) := \frac{\sum_{\mathbf{x}_{P \setminus R}} \prod_{a \in F_{P \setminus R}} \prod_{(I,J) \in N(P,R)} m_{I \to J}(x_J)}{\prod_{(I,J) \in D(P,R)} m_{I \to J}(x_J)} \tag{2.37}$$

where $N(P, R)$ is the set of all connected pairs of regions $(I, J)$ such that $J$ is in $\mathcal{E}(P)$ but not $\mathcal{E}(R)$ while $I$ is *not* in $\mathcal{E}(P)$. $D(P, R)$ is the set of all connected pairs of regions $(I, J)$ such that $J$ is in $\mathcal{E}(R)$, while $I$ is in $\mathcal{E}(P)$ , but not $\mathcal{E}(R)$.

**Example**

We consider the region graph in Figure 2.4. The belief $b_R(\mathbf{x}_R)$ at region $R$ is the product of its local factors $\prod_{a \in R} f_a(\mathbf{x}_a)$, the messages from its parents $m_{A \to R}(\mathbf{x}_R)$, and $m_{B \to R}(\mathbf{x}_R)$, and the messages into descendants from other parents who are not descendants: $m_{C \to E}(\mathbf{x}_E)$, $m_{C \to H}(\mathbf{x}_H)$ and $m_{F \to H}(\mathbf{x}_H)$.

The belief at region $R$ is given by

$$b_R \propto m_{A \to R} m_{B \to R} m_{C \to E} m_{C \to H} m_{F \to H} \prod_{a \in R} f_a(\mathbf{x}_a)$$

Similarly, for the message update from parent $R$ to child $E$, the sets $N(R, E) = \{(A, R), (B, R)\}$ and $D(R, E) = \{(D, G)\}$. The message update equation is given by

$$m_{R \to E}(\mathbf{x}_E) := \frac{\sum_{\mathbf{x}_R \setminus \mathbf{x}_E} m_{A \to R}(\mathbf{x}_R) m_{B \to R}(\mathbf{x}_R) \prod_{a \in R \setminus E} f_a(\mathbf{x}_a)}{m_{D \to G}(\mathbf{x}_G)}$$

## 2.5 Problem Definition

We have introduced the Cluster Variation Method and the conceptually similar Generalized Belief Propagation algorithm . We note that the techniques described above are applicable to the static inference case, i.e., when there is no temporal evolution in the probabilities, and a single run of the algorithm provides the approximate beliefs.

However, we are interested in the problem of tracking the evolution of marginal beliefs , not just their steady states. We now present the problem of inference on distributions evolving over time.

We consider a system of $N$ discrete-valued random processes $\mathbf{X}(t) = \{X_0(t), \ldots, X_{N-1}(t)\}$ having the Markov property, i.e., for times $0 \leq t_1 \leq \ldots \leq t_{m-1} \leq t_m$, we have $P(\mathbf{x}^{t_m} | \mathbf{x}^{t_{m-1}}, \ldots, \mathbf{x}^{t_1}) = P(\mathbf{x}^{t_m} | \mathbf{x}^{t_{m-1}})$, where $\mathbf{x}^t$ stands for the system having state $\{x_0, \ldots, x_{N-1}\}$ at time $t$ and $P(\mathbf{x}^t)$ represents the probability of the system having state $\mathbf{x}^t$ at time $t$, i.e., $P(\mathbf{X}(t) = \mathbf{x}^t)$.

Given a conditional probability distribution $\hat{p}(\mathbf{x}^{t+\delta t} | \mathbf{x}^t)$ which indicates the probability that the system has state $\mathbf{x}^{t+\delta t}$ at time $(t + \delta t)$, given that it had state $\mathbf{x}^t$ which can be factorized into localized interactions, denoted using the factor graph notation:

$$p(\mathbf{x}^{t+\delta t} | \mathbf{x}^t) = \frac{1}{Z(\mathbf{x}^t)} \prod_{a \in A} f_a(\mathbf{x}^{t+\delta t}{}_a | \mathbf{x}^t{}_a) = \frac{1}{Z(\mathbf{x}^t)} \exp\{-H(\mathbf{x}^{t+\delta t} | \mathbf{x}^t)\} \quad (2.38)$$

where $H(\mathbf{x}^{t+\delta t} | \mathbf{x}^t) = -\sum_{a \in A} \ln f_a(\mathbf{x}^{t+\delta t}{}_a | \mathbf{x}^t{}_a)$ and the conditional partition function is $Z(\mathbf{x}^t) = \sum_{\mathbf{x}^t} \prod_{a \in A} f_a(\mathbf{x}^{t+\delta t}{}_a | \mathbf{x}^t{}_a)$. Given an initial trial probability distribution $\{b(\mathbf{x}^t)\}$ at time $t = 0$, we wish to find $\{b(\mathbf{x}^t)\} \ \forall t \geq 0$.

## 2.6 Extended GBP formulation

We now extend the GBP algorithm for the problem mentioned above. Let $\mathcal{R}_{static} = \{(\alpha_s, c_{\alpha_s})\}$ be a valid region graph for the static GBP case, satisfying (2.34). We consider the discrete time case, where the "space-time" factor graph is given by

$$p(\mathbf{x}^{t+1}|\mathbf{x}^t) = \frac{1}{Z(\mathbf{x}^t)} \prod_{a \in A} f_a(\mathbf{x}_a^{t+1}|\mathbf{x}_a^t) = \frac{1}{Z(\mathbf{x}^t)} \exp\{-H(\mathbf{x}^{t+1}|\mathbf{x}^t)\} \quad t \in \{1, \ldots, T\} \quad (2.39)$$

and initial probability distribution $\{b(\mathbf{x}^t)\}$ at time $t = 0$ is given.

The set of regions, $R_{dyn}$ for the extended GBP algorithm consist of the following:

- The joint beliefs at time $t$ and $t-1$, $\alpha^{t,t-1} = \{\alpha_s^t, \alpha_s^{t-1}\}$ with counting number $c_{\alpha^{t,t-1}} = c_{\alpha_s}$ and factors $\prod_{a \in \alpha_s} f_a(\mathbf{x}_a^{t+1}|\mathbf{x}_a^t)$ .

- The single state beliefs at time $t$, given by $\alpha^t = \alpha_s^t$ with counting number $c_{\alpha^t} = -c_{\alpha_s}$ an no factors except initial at time $t = 0$ which will have factor $b(\mathbf{x}_{\alpha_s}^0)$.

Then, each factor $a$ at for the time $t|t-1$ is part of one joint region $\alpha^{t,t-1}$, while each variable node $i$ at time $t$ is part of three regions: one past-present evolution region $\alpha^{t,t-1}$, one present to next time evolution region $\alpha^{t+1,t}$ and one for current beliefs at time $t$, $\alpha^t$.

The counting numbers for a variable $i$, and factor $a$ at time $t$ are given as:

$$\sum_{\alpha \ni a} c_{\alpha^{t,t-1}} = \sum_{\alpha_s \in a} c_{\alpha_s} = 1 \quad \forall a \quad \text{(factor)} \quad (2.40)$$

$$\sum_{\alpha^{t,t-1} \ni i} c_{\alpha^{t,t-1}} + \sum_{\alpha^{t+1,t} \ni i} c_{\alpha^{t+1,t}} + \sum_{\alpha^t \ni i} c_{\alpha^t} = \sum_{\alpha_s \in a} c_{\alpha_s} + \sum_{\alpha_s \in a} c_{\alpha_s} + \sum_{\alpha_s \ni a} -c_{\alpha_s}$$
$$= 1 + 1 - 1 = 1 \quad \forall i \quad \text{(variable)} \quad (2.41)$$

which satisfy the GBP constraints (2.34).

A key requirement in inference over time is causality, i.e., the future states should not affect the past states. This is achieved due to the additional state regions $\alpha^t$, which restrict the backward flow of messages from joint regions$\alpha^{t+1,t}$ to $\alpha^{t,t-1}$. The message passing is in the forward direction, as illustrated by Figure 2.5.  This is similar to the forward filtering approach in [52].  The message update rules are influenced by the direction of time and are no longer generic.

We present an alternative formulation in the next chapter, which is naturally suited to such evolution problems.

## Summary

In this chapter, we have briefly reviewed the existing techniques of inference over graphical models.  We focus on the variational methods of approximate inference, and the relation between variational free energy and belief propagation algorithm.

The next chapter introduces the dynamic equivalent of the Cluster Variation Method known as the Path Probability Method, which is a powerful technique that allows us to track the changing beliefs.  We derive an alternative algorithm, DynBP, based on PPM, which is naturally suited to handle inference over time.

We shall revisit the GBP algorithm in Chapter 4 to consider the extension of the GBP algorithm to inference over time.

Figure 2.5: Figure presents an extended region graph for inference over time. (i) Region graph with underlying factor graph. The regions are marked as a) $\alpha^{t,t-1}$, b) $\alpha^{t,t+1}$, and i) $\alpha^0$ with associated factor $b(\mathbf{x}_\alpha^0)$. (ii) Figure shows the direction of message passing in the region graph which maintains causality. This is similar to forward filtering pass in spatio-temporal MRF of [52]

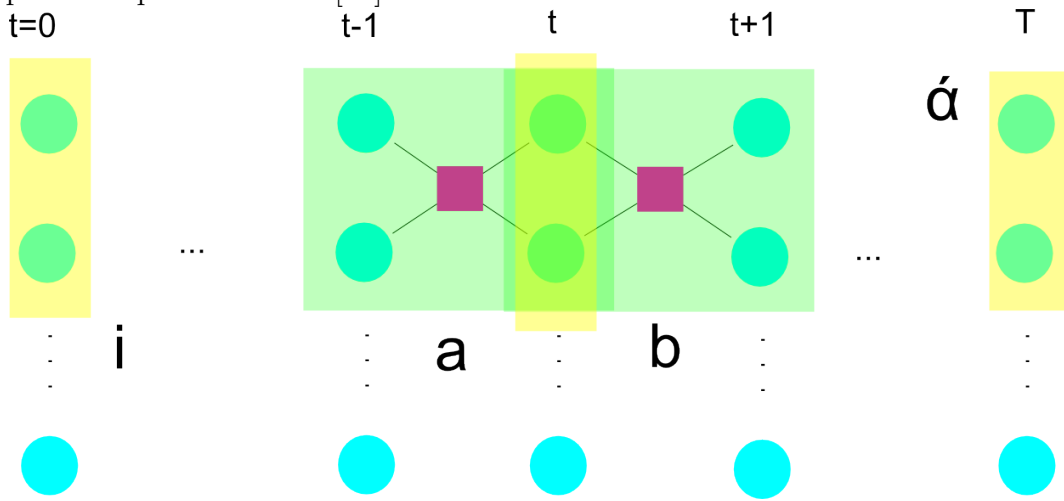

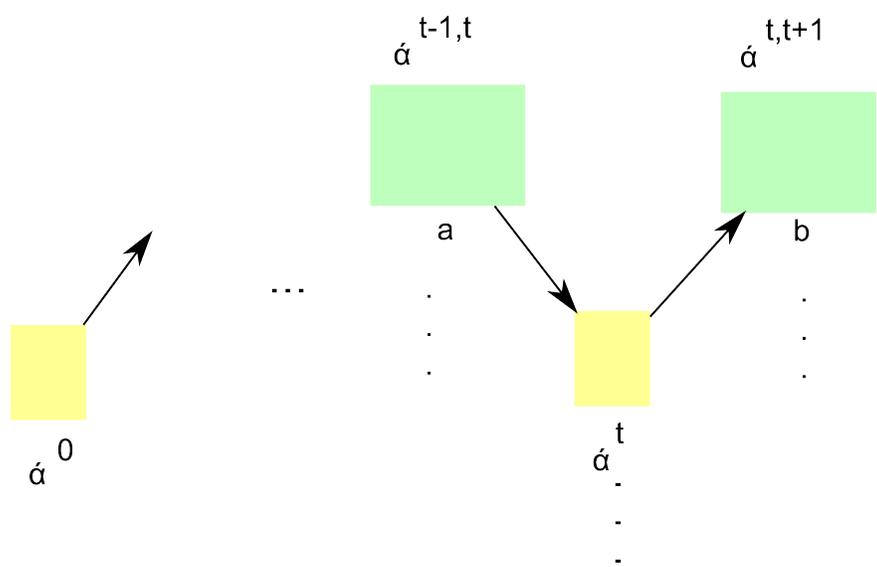

# Chapter 3

# Path Probability Method

Path probability methods (PPM) was first studied by Kikuchi [21] in the context of dynamic evolution of global ensemble quantities such as magnetism in spin glass systems. It has been successfully applied to a number of problems in statistical physics such as hopping conduction problems of many classical particles, lattice gas models, etc. A review of the theory and applications of the method in classical physics in found in [27, 33, 34]. The method is a powerful technique for studying temporal evolution of dynamical interacting multi-body systems. However, to the best of the author's knowledge, this thesis is the first attempt which extends and successfully applies this powerful technique to non-traditional problems in machine learning.

We present a brief introduction to the method as it has been developed in the context of statistical physics in section 3.1. We extend the ideas present to applications in graphical models in Section 3.2 and present the resulting algorithm, DynBP, in Section 3.3.

## 3.1    Review of the method

Path Probability Method applies the variational principle of minimum free energy to irreversible statistical mechanics. PPM defines *path variables* (analogous to the state variables in the CVM) which capture the information about the change of state of the system and constructs a *path probability function* (PPF) which expresses the probability of a change occurring in terms of the path variables. Maximization of PPF leads to the most probable path of evolution of the system.

The papers by [21] and [15] investigated a homogeneous ferromagnetic Ising system with $N$ spin sites, with the Hamiltonian energy given by

$$\mathcal{H}(\mathbf{s}) = -J \sum_{<i,j>} s_i s_j - \mu_0 H \sum_i s_i \tag{3.1}$$

where $s_i \in \{+1, -1\}$ is the Ising spin on the $i$th site, $<i,j>$ represents the nearest-neighbor pair, and $\mu_0 H$ and $J$ are constants. The probability of the system being in state $\mathbf{x}$ when in equilibrium is given by the Boltzmann distribution $P(\mathbf{x}) = \exp\{-\mathcal{H}(\mathbf{x})\}/Z$.

**State Variables**

The quantity of interest was the ensemble magnetism exhibited by the system. This depends on the total number of interacting as well as opposing spin pairs in the system. Thus, the state variables quantify the magnetism of the system under equilibrium, $\mathcal{X}_s = \{x_i, y_{ij}\},\ i,j \in \{\pm 1\}$, are the probabilities of finding a given spin state or pair interaction state at a randomly selected spin sites as given by Table 3.1, where $z$ is the lattice number of the system, i.e., $zN$ is the total number of spin pairs in the system. The state variables are inter-related probabilisties with the relations given by

$$
\begin{aligned}
x_{+1} + x_{-1} = y_{+1-1} + y_{+1+1} + y_{-1+1} + y_{-1-1} &= 1 \\
y_{+1-1} + y_{+1+1} = y_{-1+1} + y_{+1+1} &= x_{+1} \\
y_{+1-1} + y_{-1-1} = y_{-1+1} + y_{-1-1} &= x_{-1}
\end{aligned}
$$

The hamiltonian, $\mathcal{H}(\mathbf{s})$, can be rewritten in terms of the ensemble state variables, $\mathcal{X}_s$, as

$$\mathcal{H}(\mathcal{X}_s) = -\frac{JzN}{2}\{(y_{+1+1} + y_{-1-1}) - (y_{+1-1} + y_{-1+1})\} - \mu_0 HN(x_{+1} - x_{-1}) \tag{3.2}$$

Table 3.1: State variables of the spin glass system, $\mathcal{X}_s$. The state variables represent the probability of finding a random spin site or spin pair in the given state.

| State Variable | Mathematical Definition | Probability of finding state |
|:---:|:---:|:---:|
| $x_{+1}$ | $\frac{1}{N}\sum_{i=1}^{N}\mathbf{1}_{s_i=+1}$ | $\oplus$ spin |
| $x_{-1}$ | $\frac{1}{N}\sum_{i=1}^{N}\mathbf{1}_{s_i=-1}$ | $\ominus$ spin |
| $y_{+1+1}$ | $\frac{1}{zN}\sum_{<i,j>}\mathbf{1}_{s_i=+1,s_j=+1}$ | $\oplus - \oplus$ pair |
| $y_{+1-1}$ | $\frac{1}{zN}\sum_{<i,j>}\mathbf{1}_{s_i=+1,s_j=-1}$ | $\oplus - \ominus$ pair |
| $y_{-1+1}$ | $\frac{1}{zN}\sum_{<i,j>}\mathbf{1}_{s_i=-1,s_j=+1}$ | $\ominus - \oplus$ pair |
| $y_{-1-1}$ | $\frac{1}{zN}\sum_{<i,j>}\mathbf{1}_{s_i=+1,s_j=+1}$ | $\ominus - \ominus$ pair |

**Path Variables**

The method studies the evolution of the state variables, $\mathcal{X}_s$, with time. It uses the notion of *path variables*, which capture the probability of change of the spins at a random site within a small time, $\delta t$. The path variables are $\mathcal{X}_p = \{X_{i,j}, Y_{ij,kl}\}$, $i,j,k,l \in \{\pm 1\}$ where $X_{i,j}$ is the probability that a node with spin $i$ at time $t$ will change to spin $j$ at time $(t + \delta t)$, and $Y_{ij,kl}$ is the probability that a bond pair having state $(i, j)$ at time $t$ will change to state $(k, l)$ at time $(t + \delta t)$. The time $\delta t$ is chosen small enough such that at most one of the states changes in $Y_{ij,kl}$.

The expression for the path variables are given in Table 3.2 where $s_i^t$ denotes the state of spin site $i$ at time $t$. The relations between state variables at time $t$, $\mathcal{X}_s^t$, and $t + \delta t$, $\mathcal{X}_s^{t+\delta t}$, and the path variables $\mathcal{X}_p$ are given as

$$x_i^{t+\delta t} - x_i^t \;=\; X_{-i,i} - X_{i,-i} \quad \forall i \in \pm 1 \tag{3.3}$$

$$y_{ij}^{t+\delta t} - y_{ij}^t \;=\; (Y_{-ij,ij} + Y_{i-j,ij}) - (Y_{ij,i-j} + Y_{ij,-ij}) \quad \forall i,j \in \pm 1 \tag{3.4}$$

The relations can be expressed in terms of independent variables as

$$Y_{ii,ii} \;=\; y_{ii}^t - 2Y_1(i) \quad \forall i \in \pm 1 \tag{3.5}$$

$$Y_{ij,ij} \;=\; y_{ij}^t - [Y_2(i) + Y_2(-i)] \quad \forall i,j \in \pm 1 \tag{3.6}$$

The change in the hamiltonian, $\Delta\mathcal{H}(\mathcal{X}_p)$, can be expressed in terms of the path variables

Table 3.2: Path variables in the spin glass system, $\mathcal{X}_p$. The path variables represent the probabilities of the indicated change in state of a random spin site or pair.

| Path Variable | Abbr. | Change in state | | Expression |
|:---:|:---:|:---:|:---:|:---:|
| $\mathcal{X}_p$ | | $s_i^t$ | $s_i^{t+\delta t}$ | |
| $X_{+1,+1}$ | | $\oplus$ | $\oplus$ | $\frac{1}{N}\sum_{i=1}^{N}\mathbf{1}_{s_i^t=+1}\mathbf{1}_{s_i^{t+\delta t}=+1}$ |
| $X_{-1,-1}$ | | $\ominus$ | $\ominus$ | $\frac{1}{N}\sum_{i=1}^{N}\mathbf{1}_{s_i^t=-1}\mathbf{1}_{s_i^{t+\delta t}=-1}$ |
| $X_{+1,-1}$ | $X(1)$ | $\oplus$ | $\ominus$ | $\frac{1}{N}\sum_{i=1}^{N}\mathbf{1}_{s_i^t=+1}\mathbf{1}_{s_i^{t+\delta t}=-1}$ |
| $X_{-1,+1}$ | $X(-1)$ | $\ominus$ | $\oplus$ | $\frac{1}{N}\sum_{i=1}^{N}\mathbf{1}_{s_i^t=-1}\mathbf{1}_{s_i^{t+\delta t}=+1}$ |
| $Y_{+1+1,+1+1}$ | | $\oplus-\oplus$ | $\oplus-\oplus$ | $\frac{1}{zN}\sum_{<i,j>}\mathbf{1}_{s_i^t=1}\mathbf{1}_{s_j^t=1}\mathbf{1}_{s_i^{t+\delta t}=1}\mathbf{1}_{s_j^{t+\delta t}=1}$ |
| $Y_{-1+1,-1+1}$ or $Y_{+1-1,+1-1}$ | | $\oplus-\ominus$ | $\oplus-\ominus$ | $\frac{1}{zN}\sum_{<i,j>}\mathbf{1}_{s_i^t=1}\mathbf{1}_{s_j^t=1}\mathbf{1}_{s_i^{t+\delta t}=-1}\mathbf{1}_{s_j^{t+\delta t}=1}$ |
| $Y_{-1-1,-1-1}$ | | $\ominus-\ominus$ | $\ominus-\ominus$ | $\frac{1}{zN}\sum_{<i,j>}\mathbf{1}_{s_i^t=-1}\mathbf{1}_{s_j^t=-1}\mathbf{1}_{s_i^{t+\delta t}=-1}\mathbf{1}_{s_j^{t+\delta t}=-1}$ |
| $Y_{+1+1,+1-1}$ or $Y_{+1+1,-1+1}$ | $Y_1(1)$ | $\oplus-\oplus$ | $\oplus-\ominus$ | $\frac{1}{zN}\sum_{<i,j>}\mathbf{1}_{s_i^t=1}\mathbf{1}_{s_j^t=1}\mathbf{1}_{s_i^{t+\delta t}=1}\mathbf{1}_{s_j^{t+\delta t}=-1}$ |
| $Y_{-1-1,+1-1}$ or $Y_{-1-1,-1+1}$ | $Y_1(-1)$ | $\ominus-\ominus$ | $\ominus-\oplus$ | $\frac{1}{zN}\sum_{<i,j>}\mathbf{1}_{s_i^t=-1}\mathbf{1}_{s_j^t=-1}\mathbf{1}_{s_i^{t+\delta t}=-1}\mathbf{1}_{s_j^{t+\delta t}=+1}$ |
| $Y_{+1-1,+1+1}$ or $Y_{-1+1,+1+1}$ | $Y_2(1)$ | $\oplus-\ominus$ | $\oplus-\oplus$ | $\frac{1}{zN}\sum_{<i,j>}\mathbf{1}_{s_i^t=-1}\mathbf{1}_{s_j^t=1}\mathbf{1}_{s_i^{t+\delta t}=1}\mathbf{1}_{s_j^{t+\delta t}=1}$ |
| $Y_{+1-1,-1-1}$ or $Y_{-1+1,-1-1}$ | $Y_2(-1)$ | $\oplus-\ominus$ | $\ominus-\ominus$ | $\frac{1}{zN}\sum_{<i,j>}\mathbf{1}_{s_i^t=-1}\mathbf{1}_{s_j^t=1}\mathbf{1}_{s_i^{t+\delta t}=-1}\mathbf{1}_{s_j^{t+\delta t}=-1}$ |

as

$$
\begin{aligned}
\Delta\mathcal{H}(\mathcal{X}_p) &= \mathcal{H}(\mathcal{X}_s^{t+\delta t}) - \mathcal{H}(\mathcal{X}_s^t) \\
&= \sum_{i=\pm1} 2JzN[Y_1(i) - Y_2(i)] + 2\mu_0 HN[X(1) - X(-1)] \qquad (3.7)
\end{aligned}
$$

**Path Probability Function**

The probability of the path taken by the path variables, $\mathcal{X}_p$ from initial state, $\mathcal{X}_s^t$ to final state, $\mathcal{X}_s^{t+\delta t}$, is given by the *path probability function* (PPF), analogous to the free energy in the Cluster Variation Method. The method constructs an approximation to the path probability function in terms of the path variables, in which the energy change is exact, but the entropy of change in approximated. Maximizing this variational formulation leads to the most probable path of change given the initial state.

The path probability function, $\mathcal{P}$, depends on three factors, namely,

- The probability of occurrence of spin flip at a spin site in the time $\delta t$, which is given by

$$
\mathcal{P}_1 = \prod_{i=\pm1} (\theta\delta t)^{X_{i,-i}}(1-\theta\delta t)^{X_{i,i}} \qquad (3.8)
$$

where $\theta$ is the *spin-flip rate*.

- The likelihood of the change in state given by path variables, $\mathcal{X}_p$, which depends on the change in the hamiltonian, $\Delta\mathcal{H}(\mathcal{X}_p)$, in (3.7) given by

$$\mathcal{P}_2 = \exp\left(-\frac{\Delta\mathcal{H}(\mathcal{X}_p)}{k_B T}\right) \tag{3.9}$$

  where $k_B$ is the Boltzmann factor and $T$ is the temperature.

- The third factor is computed by observing the geometric relations present in the lattice. It is equivalent to the entropy in the cluster variation method, which in this case, is approximated in terms of the path variables.

$$\mathcal{P}_3 = \left[\frac{P_{\text{point}}}{P_{\text{pair}}}\right]^{z/2}\left[\frac{P_{\text{point}}}{N!}\right]^{z/2-1} \tag{3.10}$$

  where

$$
\begin{aligned}
P_{\text{point}} &= \prod_{i,j}[(NX_{i,j})!] \\
P_{\text{pair}} &= \prod_{(i,j),(i',j')}[(NY_{ij,i'j'})!]
\end{aligned}
$$

The final form of the path probability function [15] is given as:

$$
\begin{aligned}
\frac{1}{N}\ln\mathcal{P} &= (z-1)\sum_{i,j}\mathcal{L}(X_{i,j}) - \frac{z}{2}\sum_{(i,j),(k,l)}\mathcal{L}(Y_{ij,kl}) \\
&\quad + \sum_{i=\pm 1}[X_{i,-i}\ln(\theta\delta t) + X_{i,i}\ln(1-\theta\delta t) - zK(Y_1(i) - Y_2(i)) - LiX(i)]
\end{aligned} \tag{3.11}
$$

where $K = \frac{J}{k_B T}$, $L = \frac{\mu_0 H}{k_B T}$ and $\mathcal{L}(x) = x\ln x - x$. The most probable path of evolution v is obtained by differentiating $\mathcal{P}$ in (3.11) w.r.t. the independent path variables $Y_s(i)$.

We note that the third factor in (3.10) is computed using counting arguments, and thus, is heavily dependent on the lattice structure. Further, larger graphs may have local variations in the state probabilities. We extend the ideas of path probability method using the factor graph formalism to a general setting which addresses these issues. We revisit the ising spin system in A in the context of the extended setting.

## 3.2 Extension to graphical models

We consider a system of $N$ discrete-valued random processes $\mathbf{X}(t) = \{X_0(t), \ldots, X_{N-1}(t)\}$ having the Markov property, i.e., for times $0 \leq t_1 \leq \ldots \leq t_{m-1} \leq t_m$, we have $P(\mathbf{x}^{t_m}|\mathbf{x}^{t_{m-1}}, \ldots, \mathbf{x}^{t_1}) = P(\mathbf{x}^{t_m}|\mathbf{x}^{t_{m-1}})$, where $\mathbf{x}^t$ stands for the system having state $\{x_0, \ldots, x_{N-1}\}$ at time $t$ and $P(\mathbf{x}^t)$ represents the probability of the system having state $\mathbf{x}^t$ at time $t$, i.e., $P(\mathbf{X}(t) = \mathbf{x}^t)$.

Given a conditional probability distribution $\hat{p}(\mathbf{x}^{t+\delta t}|\mathbf{x}^t)$ which indicates the probability that the system has state $\mathbf{x}^{t+\delta t}$ at time $(t + \delta t)$ and had state $\mathbf{x}^t$ at time $t$, the conditional probability distribution can be factorized into localized interactions, denoted using the factor graph notation:

$$\hat{p}(\mathbf{x}^{t+\delta t}|\mathbf{x}^t) = \frac{1}{Z(\mathbf{x}^t)} \prod_{a \in A} f_a(\mathbf{x}^{t+\delta t}{}_a|\mathbf{x}^t{}_a) = \frac{1}{Z(\mathbf{x}^t)} \exp\{-H(\mathbf{x}^{t+\delta t}|\mathbf{x}^t)\} \tag{3.12}$$

where $H(\mathbf{x}^{t+\delta t}|\mathbf{x}^t) = -\sum_{a \in A} \ln f_a(\mathbf{x}^{t+\delta t}{}_a|\mathbf{x}^t{}_a)$ and the conditional partition function is $Z(\mathbf{x}^t) = \sum_{\mathbf{x}^t} \prod_{a \in A} f_a(\mathbf{x}^{t+\delta t}{}_a|\mathbf{x}^t{}_a)$. Given an initial trial probability distribution $\{b(\mathbf{x}^t)\}$ at time $t = 0$, we wish to find $\{b(\mathbf{x}^t)\} \; \forall t \geq 0$.

We use PPM to get a two-step iterative solution for finding $\{b(\mathbf{x}^t)\} \; \forall t \geq 0$. The first step is to compute an approximate joint distribution $\{b(\mathbf{x}^{t,\delta t})\}$ where $\mathbf{x}^{t,\delta t}$ denotes the joint state $(\mathbf{x}^{t+\delta t}, \mathbf{x}^t)$ which means that the system has state $\mathbf{x}^t$ at time $t$ and $\mathbf{x}^{t+\delta t}$ at time $(t + \delta t)$, subject to the constraint

$$\sum_{\mathbf{x}^{t+\delta t}} b(\mathbf{x}^{t,\delta t}) = b(\mathbf{x}^t) \tag{3.13}$$

where the criterion is minimization of the Kullback-Liebler divergence between $\{b(\mathbf{x}^{t,\delta t})\}$

and the probability distribution $\{\tilde{p}(\mathbf{x}^{t,\delta t})\}$ obtained by the relation

$$\tilde{p}(\mathbf{x}^{t,\delta t}) = \hat{p}(\mathbf{x}^{t+\delta t}|\mathbf{x}^t)b(\mathbf{x}^t).$$

The motivation is to express this minimization in terms of the beliefs $\{b_\alpha(\mathbf{x}^{t,\delta t}{}_\alpha)\}$ over the joint state $(\mathbf{x}^{t,\delta t}{}_\alpha) = (\mathbf{x}^{t+\delta t}{}_\alpha, \mathbf{x}^t{}_\alpha)$ for a valid region graph $\{\alpha : \alpha \in \mathcal{R}\}$. The beliefs $\{b_\alpha(\mathbf{x}^{t,\delta t}{}_\alpha)\}$ correspond to the path variables in the original PPM formulation.

We note that the KL divergence between true joint probability distribution $\{p(\mathbf{x}^{t,\delta t})\}$ and $\{b(\mathbf{x}^{t,\delta t})\}$ is given as

$$\begin{aligned}
D(\{b(\mathbf{x}^{t,\delta t})\}||\{p(\mathbf{x}^{t,\delta t})\}) &= D(\{b(\mathbf{x}^{t,\delta t})\}||\{\tilde{p}(\mathbf{x}^{t,\delta t})\}) + D(\{b(\mathbf{x}^t)\}||\{p(\mathbf{x}^t)\}) & (3.14) \\
&\geq D(\{b(\mathbf{x}^{t,\delta t})\}||\{\tilde{p}(\mathbf{x}^{t,\delta t})\}) & (3.15)
\end{aligned}$$

where $p(\mathbf{x}^t)$ is the true probability distribution at time $t$ and $p(\mathbf{x}^{t,\delta t}) = p(\mathbf{x}^t)\hat{p}(\mathbf{x}^{t+\delta t}|\mathbf{x}^t)$.

The second step computes the probability distribution $\{b(\mathbf{x}^{t+\delta t})\}$ according to the relation $\sum_{\mathbf{x}^t} b(\mathbf{x}^{t,\delta t}) = b(\mathbf{x}^{t+\delta t})$. Minimization of Kullback-Liebler divergence in (3.15) leads to the formulation

$$\min_{\{b(\mathbf{x}^{t,\delta t})\}} \mathcal{F}^p(\{b\}) = H^p(\{b\}) - S^p(\{b\}) + S^o(\{b\}) \qquad (3.16)$$

where $\mathcal{F}^p(\{b(\mathbf{x}^{t,\delta t})\})$ is the logarithmic form of the *path probability function* (PPF) analogous to the variational free energy $\mathcal{F}$ in CVM, the *path conditional energy* $H^p(\{b\})$, *path entropy* $S^p(\{b\})$, and the original *state entropy* $S^o(\{b\})$ are given by

$$\begin{aligned}
H^p(\{b\}) &= \sum_{\mathbf{x}^{t,\delta t}} b(\mathbf{x}^{t,\delta t})H(\mathbf{x}^{t+\delta t}|\mathbf{x}^t) & (3.17) \\
S^p(\{b\}) &= -\sum_{\mathbf{x}^{t,\delta t}} b(\mathbf{x}^{t,\delta t})\ln b(\mathbf{x}^{t,\delta t}) & (3.18) \\
S^o(\{b\}) &= -\sum_{\mathbf{x}^{t,\delta t}} b(\mathbf{x}^{t,\delta t})\ln b(\mathbf{x}^t) & (3.19)
\end{aligned}$$

Analogous to the CVM case, the PPF $\mathcal{F}^p$ can be approximated for a valid set of maximal clusters $\mathcal{R}$ with associated counting numbers $\{c_\alpha|\ \alpha \in \mathcal{R}\}$ as

$$\mathcal{F}^p(\{b_\alpha(\mathbf{x}^{t,\delta t}{}_\alpha), \alpha \in \mathcal{R}\}) = \sum_{\alpha \in \mathcal{R}} c_\alpha \mathcal{F}^p_\alpha(\{b_\alpha(\mathbf{x}^{t,\delta t}{}_\alpha)\}) \tag{3.20}$$

where $\mathbf{x}^{t,\delta t}{}_\alpha$ denotes the joint state $(\mathbf{x}^{t+\delta t}{}_\alpha, \mathbf{x}^t{}_\alpha)$ for any cluster $\alpha$, and the cluster path probability function for cluster $\alpha$ having associated *path variables* $\{b_\alpha(\mathbf{x}^{t,\delta t}{}_\alpha)\}$ are given by

$$\mathcal{F}^p_\alpha(\{b_\alpha\}) = H^p_\alpha(\{b_\alpha\}) - S^p_\alpha(\{b_\alpha\}) + S^o_\alpha(\{b_\alpha\}) \tag{3.21}$$

and the *region average path energy* $H^p_\alpha(\{b_\alpha\})$, the *region average path entropy* $S^p_\alpha(\{b_\alpha\})$, the *region average state entropy* $S^o_\alpha(\{b_\alpha\})$ and the conditional region energy, $H_\alpha(\mathbf{x}^{t+\delta t}{}_\alpha|\mathbf{x}^t{}_\alpha)$ are given by

$$H^p_\alpha(\{b_\alpha\}) = \sum_{\mathbf{x}^{t,\delta t}{}_\alpha} b_\alpha(\mathbf{x}^{t,\delta t}{}_\alpha) \tag{3.22}$$

$$H_\alpha(\mathbf{x}^{t+\delta t}{}_\alpha|\mathbf{x}^t{}_\alpha) = -\sum_{a \in \alpha} \ln f_a(\mathbf{x}^{t+\delta t}{}_a|\mathbf{x}^t{}_a) \tag{3.23}$$

$$S^p_\alpha(\{b_\alpha\}) = -\sum_{\mathbf{x}^{t,\delta t}{}_\alpha} b_\alpha(\mathbf{x}^{t,\delta t}{}_\alpha) \ln b_\alpha(\mathbf{x}^{t,\delta t}{}_\alpha) \tag{3.24}$$

$$S^o_\alpha(\{b_\alpha\}) = -\sum_{\mathbf{x}^{t,\delta t}{}_\alpha} b_\alpha(\mathbf{x}^{t,\delta t}{}_\alpha) \ln b_\alpha(\mathbf{x}^t{}_\alpha) \tag{3.25}$$

PPM reduces to minimization of the variational PPF (3.20) w.r.t. the path variables $\{b(\mathbf{x}^{t,\delta t}{}_\alpha)|\ \mathbf{x}^{t,\delta t}{}_\alpha,\ \alpha \in \mathcal{R}\}$ subject to the constraints

$$\sum_{\mathbf{x}^{t+\delta t}{}_\alpha} b_\alpha(\mathbf{x}^{t,\delta t}{}_\alpha) = b_\alpha(\mathbf{x}^t{}_\alpha) \quad \forall \alpha \in \mathcal{R} \tag{3.26}$$

$$\sum_{\mathbf{x}^{t,\delta t}{}_{\alpha\backslash\beta}} b_\alpha(\mathbf{x}^{t,\delta t}{}_\alpha) = b_\beta(\mathbf{x}^{t,\delta t}{}_\beta) \quad \forall \beta \subset \alpha \in \mathcal{R} \tag{3.27}$$

The choice of the trial conditional probability distribution $\hat{p}(\mathbf{x}^{t+\delta t}|\mathbf{x}^t)$ is constrained as in (3.12). This is a necessary condition for the decomposition of the path conditional energy $H^p$ into cluster path energies $\{H^p_\alpha|\alpha \in \mathcal{R}\}$ for a valid region graph to be accurate.

The approximate probability distribution $b(\mathbf{x}^{t+\delta t})$ is related to the cluster probability distributions $\{b_\alpha(\mathbf{x}^{t+\delta t}{}_\alpha),\ \alpha \in \mathcal{R}\}$ as $b(\mathbf{x}^{t+\delta t}) \propto \prod_{\alpha \in \mathcal{R}} b_\alpha(\mathbf{x}^{t+\delta t}{}_\alpha)^{c_\alpha}$.

We present an iterative message passing algorithm to obtain the evolution of beliefs in the next section.

## 3.3 DynBP Algorithm

This section presents the iterative update algorithm, DynBP[12], which solves the optimization problem having cost function (3.20) subject to the constraints (3.26) and (3.27) iteratively. We get the Lagrangian $\mathcal{L}$ as

$$
\begin{aligned}
\mathcal{L} &= \sum_{\alpha \in \mathcal{R}} c_\alpha \sum_{\mathbf{x}^{t,\delta t}{}_\alpha} b_\alpha(\mathbf{x}^{t,\delta t}{}_\alpha)\{\ln b_\alpha(\mathbf{x}^{t,\delta t}{}_\alpha) + H_\alpha(\mathbf{x}^{t,\delta t}{}_\alpha) - \ln b_\alpha(\mathbf{x}^t{}_\alpha)\} \\
&+ \sum_{\alpha \in \mathcal{R}} \sum_{\mathbf{x}^t{}_\alpha} \lambda_\alpha(\mathbf{x}^t{}_\alpha)\{b_\alpha(\mathbf{x}^t{}_\alpha) - \sum_{\mathbf{x}^{t+\delta t}_\alpha} b_\alpha(\mathbf{x}^{t,\delta t}_\alpha)\} \\
&+ \sum_{\beta \subset \alpha \in \mathcal{R}} \sum_{\mathbf{x}^{t,\delta t}{}_\beta} \lambda_{\beta \rightarrow \alpha}(\mathbf{x}^{t,\delta t}{}_\beta)\{b_\beta(\mathbf{x}^{t,\delta t}{}_\beta) - \sum_{\mathbf{x}^{t,\delta t}{}_{\alpha \backslash \beta}} b_\alpha(\mathbf{x}^{t,\delta t}{}_\alpha)\}
\end{aligned}
\tag{3.28}
$$

where $\lambda_\alpha(\mathbf{x}^t{}_\alpha)$ is the lagrangian coefficient corresponding to constraint (3.26) for each state $\mathbf{x}^t{}_\alpha$ of region $\alpha$, and $\lambda_{\beta \rightarrow \alpha}(\mathbf{x}^{t,\delta t}{}_\beta)$ is the lagrangian coefficients corresponding to constraints (3.27) for each joint state $\mathbf{x}^{t,\delta t}{}_\beta$ of edge $(\alpha, \beta)$ respectively. We denote $m_{\beta \rightarrow \alpha}(\mathbf{x}^{t,\delta t}{}_\beta) = \exp\{\lambda_{\beta \rightarrow \alpha}(\mathbf{x}^{t,\delta t}{}_\beta)\}$ and $m_{\alpha \rightarrow \gamma}(\mathbf{x}^{t,\delta t}{}_\alpha) = \exp\{\lambda_{\alpha \rightarrow \gamma}(\mathbf{x}^{t,\delta t}{}_\alpha)\}$ as the messages corresponding to the child region $\beta$ and parent region $\gamma$ of region $\alpha$ respectively for each joint state $\mathbf{x}^{t,\delta t}{}_\alpha$, and $m_\alpha(\mathbf{x}^t{}_\alpha) = \exp\{\lambda_\alpha(\mathbf{x}^t{}_\alpha)\}$ as the message corresponding to the past state $\mathbf{x}^t{}_\alpha$ for cluster $\alpha$.

Differentiating $\mathcal{L}$ w.r.t. $b_\alpha(\mathbf{x}^{t,\delta t}{}_\alpha)$ and setting it to zero; and using the constraints in (3.26) and (3.27) respectively, we get an iterative update scheme for $m_{\beta \rightarrow \alpha}(\mathbf{x}^{t,\delta t}{}_\beta)$ and $m_\alpha(\mathbf{x}^t{}_\alpha)$ and $b_\alpha(\mathbf{x}^{t,\delta t}{}_\alpha)$ as: constraints in (3.26) and (3.27) respectively as:

$$
m_\alpha^{(i+1)}(\mathbf{x}^t{}_\alpha) = m_\alpha^{(i)}(\mathbf{x}^t{}_\alpha)\Big\{\frac{b_\alpha(\mathbf{x}^t{}_\alpha)}{\sum_{\mathbf{x}^{t+\delta t}{}_\alpha} b_\alpha^{(i)}(\mathbf{x}^{t,\delta t}{}_\alpha)}\Big\}^{c_\alpha}
\tag{3.29}
$$

$$
m_{\beta \rightarrow \alpha}^{(i+1)}(\mathbf{x}^{t,\delta t}{}_\beta) = m_{\beta \rightarrow \alpha}^{(i)}(\mathbf{x}^{t,\delta t}{}_\beta)\Big\{\frac{b_\beta^{(i)}(\mathbf{x}^{t,\delta t}{}_\beta)}{\sum_{\mathbf{x}^{t,\delta t}{}_{\alpha \backslash \beta}} b_\alpha^{(i)}(\mathbf{x}^{t,\delta t}{}_\alpha)}\Big\}^{\frac{c_\alpha c_\beta}{c_\alpha + c_\beta}}
\tag{3.30}
$$

---

[1] Derivation of the DynBP algorithm is given in the appendix.
[2] We call the algorithm DynBP indicating BP over time for dynamic models.

$$b_\alpha^{(i+1)}(\mathbf{x}^{t,\delta t}{}_\alpha) = \frac{b_\alpha(\mathbf{x}^t{}_\alpha)\prod_{a\in\alpha}f_a(\mathbf{x}^{t+\delta t}{}_a|\mathbf{x}^t{}_a)}{m_\alpha^{(i+1)}(\mathbf{x}^t{}_\alpha)^{-1/c_\alpha}\times e}\left\{\frac{\prod_{\beta\subset\alpha}m_{\beta\to\alpha}^{(i+1)}(\mathbf{x}^{t,\delta t}{}_\beta)}{\prod_{\alpha\subset\gamma}m_{\alpha\to\gamma}^{(i+1)}(\mathbf{x}^{t,\delta t}{}_\alpha)}\right\}^{1/c_\alpha} \quad (3.31)$$

where $i$ denotes the iteration number.

The DynBP algorithm passes messages from the child to the parent for each child joint state. The stopping criterion we chose is based on the maximum change in region beliefs compared with the previous iteration beliefs. DynBP allows handling of complex models, while maintaining temporal causality.

The final algorithm is given in 1

---

**Algorithm 1** DynBP

---

**input:** $\hat{p}(\mathbf{x}^{t+\delta t}|\mathbf{x}^t)$ on factor graph $G=(V,E)$ as in (3.12); Maximal set of clusters $\{(\alpha,c_\alpha),\ \alpha\in\mathcal{R}\}$; initial probability distribution $\{b(\mathbf{x}^t)\}$ at $t=0$
**output:** partial beliefs $\{b_\alpha(\mathbf{x}^t{}_\alpha)\}$, $\alpha\in\mathcal{R}$ at times $t_1<t_2<\cdots<t_{max}$
**init:** $t\leftarrow 0$; $m_\alpha(\mathbf{x}^t{}_\alpha)=1\ \forall\alpha\in\mathcal{R}$; $m_{\beta\to\alpha}(\mathbf{x}^{t,\delta t}{}_\beta)=1\ \forall\beta\subset\alpha\in\mathcal{R}$
**repeat**
  **for all** $\alpha\in\mathcal{R}$ **do**
    **for all** $\mathbf{x}^{t,\delta t}{}_\alpha$ **do**
      Update $b_\alpha(\mathbf{x}^{t,\delta t}{}_\alpha)$ as in (B.6)
      **for all** $\gamma\in\mathcal{R}$ such that $\alpha\subset\gamma$ **do**
        Update $m_{\alpha\to\gamma}(\mathbf{x}^{t,\delta t}{}_\alpha)$ as in (3.30)
      **end for**
    **end for**
    **for all** $\mathbf{x}^t{}_\alpha$ **do**
      Update $m_\alpha(\mathbf{x}^t{}_\alpha)$ as in (3.29)
    **end for**
  **end for**
  **if** Stopping Criterion **then**
    Set $b_\alpha(\mathbf{x}^{t+\delta t}{}_\alpha)\leftarrow\sum_{\mathbf{x}^t{}_\alpha}b_\alpha(\mathbf{x}^{t,\delta t}{}_\alpha)$
    Update $t\leftarrow t+\delta t$
    Reset $m_{\beta\to\alpha}(\mathbf{x}^{t,\delta t}{}_\beta)=1\ \forall\beta\subset\alpha\in\mathcal{R}$
  **end if**
**until** $t>t_{max}$

---

# Summary

In this chapter, we have reviewed the Path Probability Method as it appears in statistical mechanics. We have extended the Path Probability Method to approximate inference over general graphical models. This has resulted in the *DynBP* algorithm.

In the following chapter, we compare the DynBP algorithm with an implementation of GBP over the "space-time" factor graph. We also perform experiments to verify the accuracy and efficiency of the algorithm.

# Chapter 4

# Algorithm Analysis

We analyse the DynBP algorithm in this chapter and compare it against existing approaches. Section 4.1 explores the relation between DynBP and GBP algorithm. Section 4.2 presents the experiments for accuracy and efficiency of the algorithm.

## 4.1 Comparison of DynBP with GBP

DynBP is closely related to GBP. Standard GBP is a static algorithm, i.e., there is no concept of temporal evolution. In this chapter, we focus on the extension of GBP to handle temporal evolution of beliefs and its relation with DynBP.

However, in cases where messages have to be passed only in the forward direction in time, DynBP is better suited to GBP. To illustrate, we consider an augmented "space-time" factor graph (3.12) with additional factors $\{b_\alpha(\mathbf{x}^t{}_\alpha)|\alpha \in \mathcal{R}\}$ given by

$$P(\mathbf{x}^{t+\delta t}, \mathbf{x}^t) \propto \prod_{a \in A} b_a(\mathbf{x}^t{}_a) \prod_{a \in A} f_a(\mathbf{x}^{t+\delta t}{}_a | \mathbf{x}^t{}_a) \quad \vec{x}_a \subset \vec{x} \tag{4.1}$$

where $P(\mathbf{x}^t) \propto \prod_{a \in A} b_a(\mathbf{x}^t{}_a)$ corresponds to the initial probability distribution at time $t$. Further, for any cluster $\alpha$, the initial cluster probability distribution $\{P(\mathbf{x}^t{}_\alpha)\}$ is given by $P(\mathbf{x}^t{}_\alpha) \propto \prod_{a \in \alpha} b_a(\mathbf{x}^t{}_a)$. The GBP algorithm yields the original cost function (3.20) subject to the original compatibility constraints (3.27) and a modified normalization

constraint

$$\sum_{\mathbf{x}^{t,\delta t}{}_\alpha} b_\alpha(\mathbf{x}^{t,\delta t}{}_\alpha) = 1 \tag{4.2}$$

instead of the original normalization constraint (3.26). Proceeding as in the previous section, We obtain the update equations as:

$$m_\alpha^{(i+1)} = m_\alpha^{(i)} \Big\{ \frac{1}{\sum_{\mathbf{x}^{t,\delta t}{}_\alpha} b_\alpha^{(i)}(\mathbf{x}^{t,\delta t}{}_\alpha)} \Big\}^{c_\alpha} \tag{4.3}$$

$$b_\alpha^{(i+1)}(\mathbf{x}^{t,\delta t}{}_\alpha) \propto \frac{b_\alpha(\mathbf{x}^t{}_\alpha) \prod_{a \in \alpha} f_a(\mathbf{x}^{t+\delta t}{}_a | \mathbf{x}^t{}_a)}{m_\alpha^{(i+1)^{-1/c_\alpha}} \times e} \Big\{ \frac{\prod_{\beta \subset \alpha} m_{\beta \to \alpha}^{(i+1)}(\mathbf{x}^{t,\delta t}{}_\beta)}{\prod_{\alpha \subset \gamma} m_{\alpha \to \gamma}^{(i+1)}(\mathbf{x}^{t,\delta t}{}_\alpha)} \Big\}^{1/c_\alpha} \tag{4.4}$$

and $m_{\beta \to \alpha}^{(i+1)}(\mathbf{x}^{t,\delta t}{}_\beta)$ same as in (3.30). The GBP algorithm does not enforce (3.26), and hence, the prior distribution $\hat{b}_\alpha^{GBP}(\mathbf{x}^t{}_\alpha)$ given by:

$$\hat{b}_\alpha^{GBP}(\mathbf{x}^t{}_\alpha) = \sum_{\mathbf{x}^{t+\delta t}{}_\alpha} b_\alpha^{GBP}(\mathbf{x}^{t,\delta t}{}_\alpha) \tag{4.5}$$

may differ from $b_\alpha(\mathbf{x}^t{}_\alpha)$, while this equality is strictly enforced by the DynBP algorithm at the cost of additional messages $m_\alpha(\mathbf{x}^t{}_\alpha)$, one for each state $\mathbf{x}^t{}_\alpha$, instead of a single message $m_\alpha$ for each cluster $\alpha$. A particular advantage of DynBP over GBP would be when strict temporal coherence is needed, i.e., the messages from future should not affect beliefs at previous nodes. DynBP provides a natural way of implementing this scheme.

We compare an implementation of DynBP with standard GBP run over the modified space-time graph and show the results in figure 4.1. The results indicate that the algorithms are comparable with the additional message requirement in case of the GBP algorithm.

## 4.2 Ising Spin Experiment

We consider the square lattice Ising model, which has $N$ binary variables arranged in an $L \times L$ square lattice, and each variable node is connected to its nearest neighbors by a pairwise factor of the form $f_a(x_i, x_j) = \exp\{J_{ij} x_i x_j\}$ and has a "local magnetic field" of the form, $f_i(x_i) = \exp\{h_i x_i\}$, where $x_i \in \{\pm 1\}$ and the parameters $J_{ij}$, $h_i$ are chosen from Gaussian probability distributions with mean 0 and variance 0.1. We propose to solve the inference problem on the graph by the DynBP algorithm.

Figure 4.1: Ratio of the variational free energy $\mathcal{F} = -\ln Z$ reported by DynBP and standard GBP implementation for N=200 trials on an Ising grid.

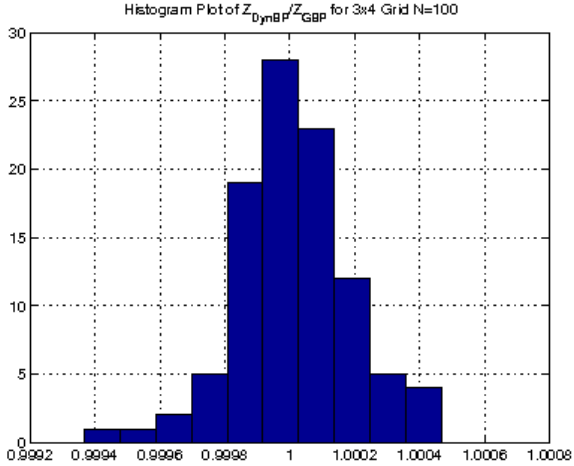

We compare the DynBP marginal probabilities at each node w.r.t. the true marginal probabilities and find that the DynBP tracks the true marginal probabilities with a high degree of accuracy, usually, with nearly 50% cases having relative absolute error within 10%, as seen from Figure 4.3.

Figure 4.2 shows the evolution of beliefs at a random node at different values of $\theta \delta t$, compared against the results obtained by loopy BP and the true evolution of belief obtained by exhaustive simulation. We note that at higher $\theta \delta t$, the spin is highly disposed to flip sign, and hence we see the oscillatory behaviour in node marginal beliefs initially.

Figure 4.4 indicates that the belief compatibility between the parent and child regions is obtained within some iterations. It can be seen that the beliefs between child and parent regions reach compatibility within a few iterations, except for some residual error in some nodes.

Figure 4.2: The figure shows the evolution of beliefs at values of $\theta \delta t$ being (a) 0.1, (b) 0.5 and (c) 0.9 respectively. We note the oscillatory behaviour of the node beliefs reported by the algorithm due to high spin flip rate. (d) Evolution of belief at various spin flip rates at the same node. Lower values of $\theta \delta t$ discourages rapid change of state. Note that at $\theta \delta t = 0.5$ the beliefs reach equilibrium in one iteration as the time factor is same for all states.

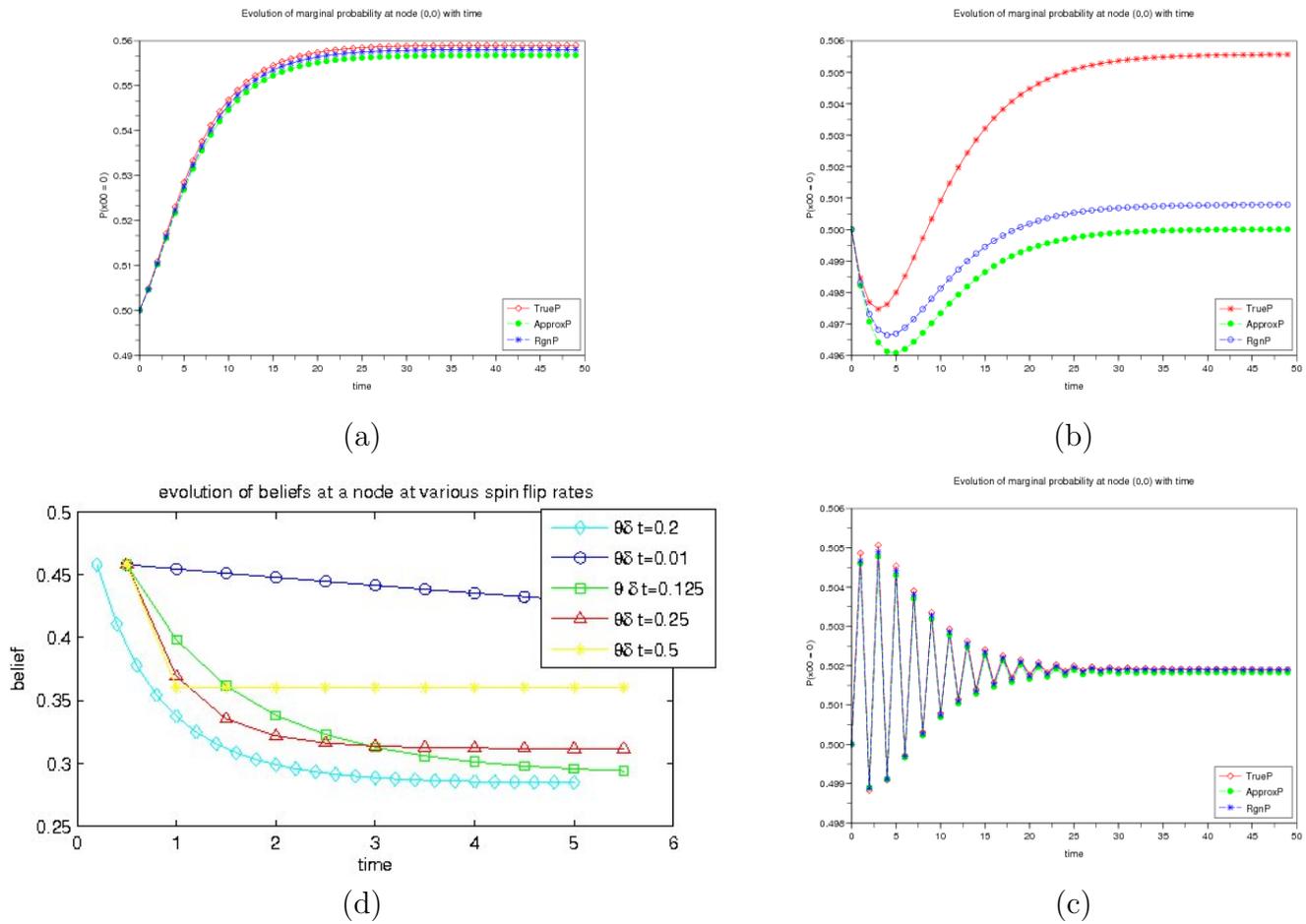

(a)                                                                      (b)

(d)                                                                      (c)

Figure 4.3: Histogram of ratio of deviation between reported beliefs and the true marginal probability against the true marginal probability at the node, $\frac{|b-p|}{p}$, at node $(0,0)$ for various seeds averaged over the simulation period for a $3 \times 4$ Ising model at various field strengths: (a) $h = 0.1$, $j = 0.5$ (b) $h = 1.0$, $j = 0.1$, (c) $h = 0.1$, $j = 0.1$.

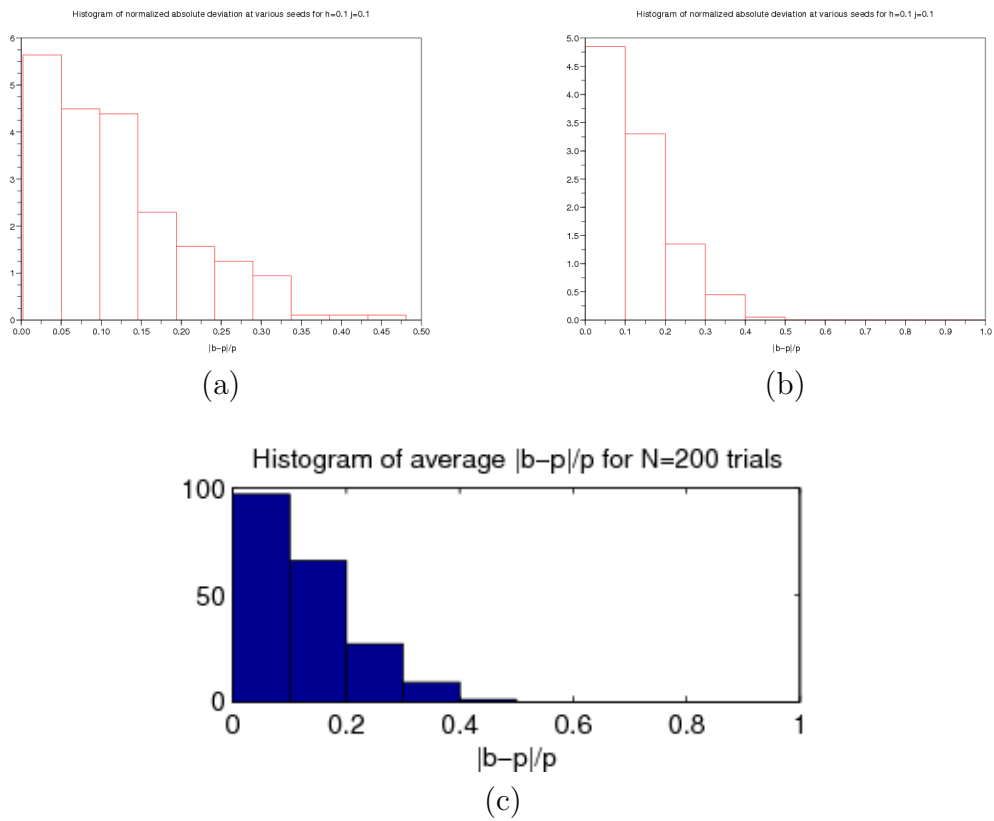

Figure 4.4:  Maximum difference between state belief for child region and corresponding marginalization in parent region; and maximum change in region beliefs with each iteration.

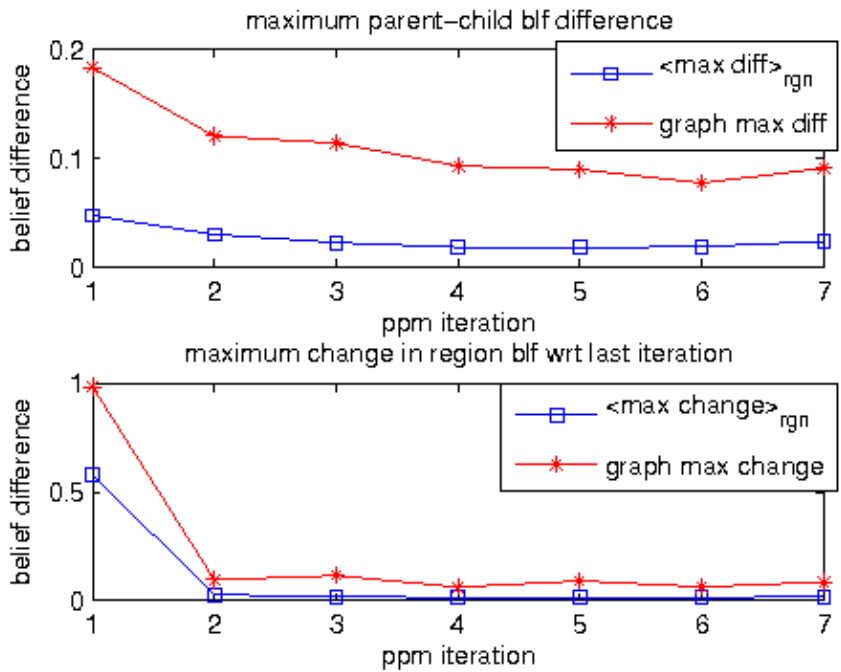

# Summary

In this chapter, we have compared the DynBP algorithm with an equivalent GBP implementation. We have also verified the accuracy of the algorithm. The next chapter presents the application of the algorithm to some computer vision problems which exhibit spatio-temporal coherence structure that can be exploited using the algorithm.

# Chapter 5

# Applications

This chapter presents the experiments and results. Section 5.1 presents the experiments on moving object detection. Section 5.2 presents the dropped frame reconstruction experiments. Section 5.3 presents the experiments on video denoising.

## 5.1 Moving Object Detection

Detection of motion in video has been a key problem in computer vision. Frame difference techniques have been widely used to detect motion based on change in pixel values [6, 14, 39]. These techniques detect motion easily but often do detect the partial outline of the object under motion rather than the complete object. Other approaches include motion history image (MHI) [3] which uses a temporal template to detect human motion, optical flow based [46], background model based history maps [13].

Yin and Collins [52] first used the idea of spatio-temporal MRFs for detecting motion in video. In this thesis, we extend their work to a formal setting under the DynBP framework.

Under the model, each pixel $(m, n)$ at time instant $k$ has a hidden state node $s(m, n, k)$ which represents the likelihood that a pixel contains object motion, and a corresponding data node $d(m, n, k)$, which represents the binary motion detection result computed by inter-frame differencing at time $k$. The network consists of nearest neighbor nodes in space and time, i.e., a node $s(m, n, k)$ has four spatial neighbors $s(m \pm 1, n, k)$ and $s(m, n \pm 1, k)$ and two temporal neighbors $s(m, n, k \pm 1)$. The joint probability over the 3D volume is given as $P(s_1, \cdots, s_N, d_1, \cdots, d_N) = \prod_{i \neq j} \psi_{ij}(s_i, s_j) \prod_k \phi_k(s_k, d_k)$ where $s_i$ and $d_i$ represent the state node and the data node separately. $\phi(s_j, d_j)$ represents the measurement relation between observation $d_j$ and hidden state node $s_j$. If $d_j$ equals zero, no motion is detected at the pixel by inter-frame differencing. The measurement relation $\phi(s_j, d_j); \forall s_j \in \{1, \cdots, C\}$ is given by $\phi_j(s_j = p, d_j) = \frac{1}{C}$ if $d_j = 0$ else $\phi_j(s_j = p, d_j) = \delta(p = C)$ where $C$ is the number of quantization bins for the belief at state node. The state transition function $\psi_i j(s_i, s_j)$ is given as $\psi_{ij}(s_i = p, s_j = q; \theta) = \theta$ if $p = q$ else $\psi_{ij}(s_i = p, s_j = q; \theta) = \epsilon$ where $\theta$ and $\epsilon$ are related as $0 < \theta < 1, \epsilon = (1 - \theta)/(C - 1)$ and $\theta \gg \epsilon$. We define a closely related formulation of the conditional probability function $\hat{p}(\vec{s}^{t+1} | \vec{s}^t)$ at any time instant $t$ as

$$\hat{p}(\vec{s}^{t+1} | \vec{s}^t) \propto \prod_{(i,j)} f_{ij}(s_i^{t+1}, s_j^{t+1}) \prod_i f_i(s_i^{t+1} | s_i^t, d_i^{t+1}) \tag{5.1}$$

where $f_{ij}(s_i^{t+1}, s_j^{t+1})$ corresponds to the spatial compatibility function and is given by $f_{ij}(s_i, s_j) = \psi_{ij}(s_i, s_j; \theta_s)$ where state nodes $s_i$ and $s_j$ are spatial neighbors. The time compatibility is captured in $f_i(s_i^{t+1} | s_i^t, d_i^{t+1})$,

$$= \begin{cases} \delta(s_i^{t+1} = (C - 1)) & \text{if } d_i^{t+1} = 1 \\ \psi_{ij}(s_i^{t+1}, l_i^t; \theta_t) & \text{otherwise} \end{cases} \tag{5.2}$$

where we choose $l_i^t = \max(0, s_i^t - 1)$, which acts as a decay function in absence of binary motion detected by inter-frame differencing($d_i^{t+1} = 0$). We construct a valid region graph using Bethe approximation, i.e., large regions consisting of spatially neighboring $(s_i, s_j)$ and small regions $s_i$, and use MAP rule on each small region as $s_i^{MAP} = \arg \max_{s_i} b_i(s_i)$ at each time instant. This is conceptually similar to forward spatial-temporal BP reported in the original paper. Figure 3 shows the results of the algorithm applied to a $80 \times 50 \times 240$ video.

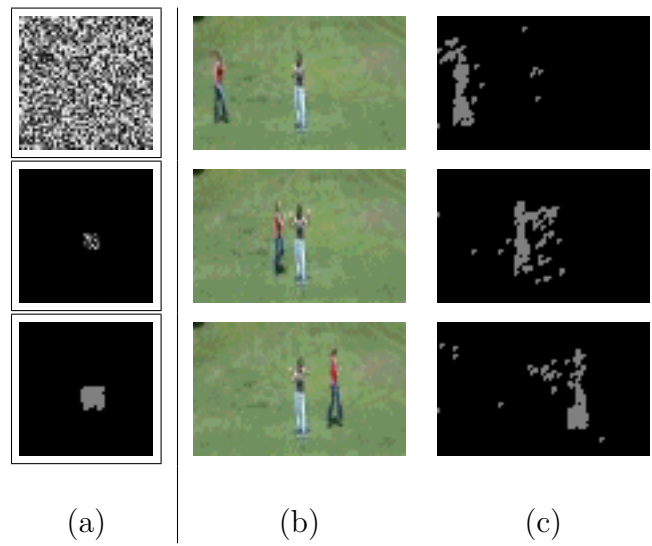

(a) (b) (c)

Figure 5.1: Moving Object Detection Experiment: *Random Patch:* (*a*) There is a random $5 \times 5$ patch moving on $50 \times 50$ background. The three vertical frames shows the image, inter-frame difference and the DynBP result respectively for the random patch.[52] report that the camouflaged object cannot be adequately detected by histogram based detector. Median and Gaussian filters do not localize the object position well. We observe that our result are similar to the forward spatio-temporal BP in [52]. *Real Video*: (*b*) The original frame and (*c*) DynBP result for frames 20, 90 & 160 for $\theta_s = 0.99$, $\theta_t = 0.6$, $C = 2$, where the input frame is quantized to 8 bins to reduce jitter.

## 5.2    Dropped frame reconstruction

Video completion and dropped frame reconstruction have gained increasing importance due to recent growth in online videos and mobile devices. A critical parameter is the duration of video frames to be filled which differs based on the task. Several approaches using incremental completion have been investigated. Wexler etal. [45] use patch based techniques to fill in the missing pixels. Jia etal. [17] use tracking and fragment merging to fill in the missing patches. Cheung etal. [5] uses probabilistic representations of the video (epitomes) to fill in the missing patches.

A key problem with applying inference algorithms to video completion is the state space size which is exponential in the number of values that can take. For any region $\alpha$, we use an approximate conditional probability which is dependent only on the number of variables $|\alpha|$, present in region $\alpha$, i.e., $\hat{p}_\alpha^{t+\delta t|t}(\mathbf{x}_\alpha^{t+\delta t}|\mathbf{x}_\alpha^t) = \hat{p}_{|\alpha|}^{t+\delta t|t}(\mathbf{x}_\alpha^{t+\delta t}|\mathbf{x}_\alpha^t)$.

A reduced search space is obtained by considering the high probability next states given the current state for each region. The search space is then augmented corresponding to the extra states which are present either in the child or parent of current region, thus obtaining a closed search space on which the algorithm can be run.

We use the following preprocessing algorithm for getting a valid search space for the PPM algorithm:

1. *PruneCandidateList():* This step selects only the candidates with high probability given the current state

2. *AugmentParent():* Add the missing states in the child region corresponding to parent region state

3. *AugmentChild():* Add the most likely states in the parent region such that all states present in the child region have equivalent representation in parent region

4. We use approximate nearest neighbour search to find the best matching candidates.

In our experiment, we consider a $300 \times 100 \times 240$ video sequence which is quantized to 8 levels, and $5 \times 5$ regions. We find that each dropped frame reconstruction takes $3 - 4$ seconds. Figure 5.2 shows the application of the PPM algorithm to te problem of dropped frame reconstruction for streaming video broadcast.

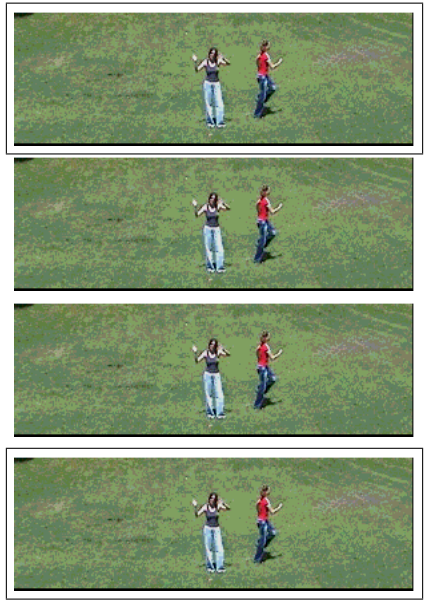

Figure 5.2: This experiment reconstructs the dropped frames when receiving streaming video over the internet. The frame-dropping is modeled as a Bernoulli process similar to [5]. The leftmost and rightmost frames are the true normalized frames 150 and 157 respectively, while the predicted frames 153 and 155 are shown in the middle.

## 5.3 Video denoising

Video denoising has become an increasingly important video processing task with the rapid growth of multimedia technology, since distortion of a video is inevitable during its acquisition, processing, storage and transmission [23].

Several video noise reduction techniques have been investigated which use spatio-temporal filters in the pixel domain. Some examples include the adaptive weighted averaging filter [32], adaptive recursive least square filter [22] and motion compensated Kalman filter [47]. Alternative methods have explored filtering in the transform domain such as wavelet transforms [37, 53, 54]. There exists a high correlation among neighbouring frames of a video, since the motion among such frames is small. This makes it well suited for application of DynBP algorithm.

Video denoising [5] is posed as an inference problem, where, some of the pixel values are missing and the missing pixel values are filled using the MAP estimate, i.e., those values which have maximum probability given the evidence posed by the known pixels. We consider a $352 \times 240$ video in which the red, green and blue components of the pixels and missing with 50% probability. A high variance noise $\sigma^2$ is added to the missing pixels. The video data which quantized to 8 levels. Figure 5.3 shows the results of applying the algorithm to the problem of video denoising.

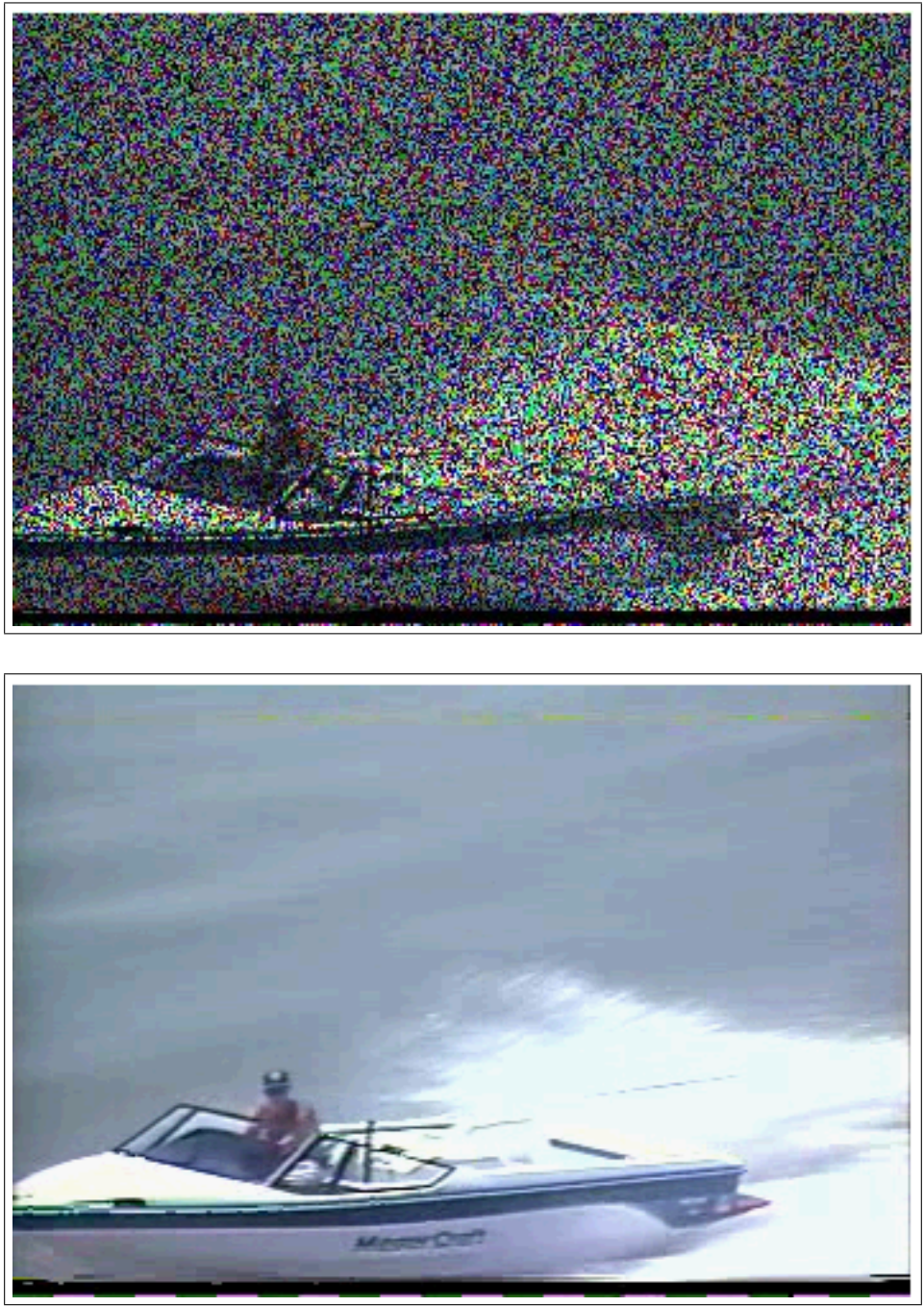

Figure 5.3: The $352 \times 240$ video has its red,green and blue pixels missing independently with 50% probability. A high variance noise is added to the missing pixels. The top frame shows the noisy video, and the bottom frame shows the result of denoising.

# Chapter 6

# Conclusions and Future Work

We have investigated the Path probability method present in statistical mechanics. We have extended it to handle approximate inference over time in graphical models. We have verified the accuracy of the algorithm by comparison against existing approaches. A special formulation based on the Generalized Belief Propagation algorithm has been shown to be equivalent to the Path Probability Method.

We have demonstrated the applicability of the algorithm to problems in computer vision which have special temporal evolution characteristics.

## 6.1   Future Work

A key point of interest is the extension of the DynBP algorithm to handle continuous time evolution. The update equations take the form of stochastic differential equations, and have to be treated differently. Tractable solutions may be obtained for some conditional evolution distributions.

Another direction of research is application of DynBP to the problem of multiple object tracking with different speeds.

# Appendix A

# Ising System Reformulation

We revisit the homogenous ferromagnetic Ising system in Section 3.1 and show how it can be recast in the extended setting. We consider a region graph based on the *Bethe-Pierls approximation,* consisting of $\mathcal{R} = \mathcal{R}_s \cup \mathcal{R}_L$, where $\mathcal{R}_S$ and $\mathcal{R}_L$ are the small and large regions. There are $N$ small regions, each consisting of a single variable node. There are $Nz/2$ large regions, each consisting of a single factor node and all the variable nodes neighbouring the factor node. The counting numbers for each region are given by

$$c_\alpha = 1 - \sum_{\beta \in \mathcal{S}(R)} c_\beta$$

where $\mathcal{S}(R)$ is the set of regions that are super-regions of $R$. Then,

$$c_\alpha = \begin{cases} 1 & \forall \alpha \in \mathcal{R}_L \\ 1 - z & \forall \alpha \in \mathcal{R}_S \end{cases} \tag{A.1}$$

The conditional region energy, $H_\alpha(\mathbf{x}^{t+\delta t}{}_\alpha | \mathbf{x}^t{}_\alpha)$, will then be given by:

$$H_\alpha(\mathbf{x}^{t+\delta t}{}_\alpha | \mathbf{x}^t{}_\alpha) = \exp\{-\mathcal{H}_\alpha(\mathbf{x}^{t+\delta t}{}_\alpha) + \mathcal{H}_\alpha(\mathbf{x}^t{}_\alpha)\} \times (\theta\delta t)^{N_\alpha^f} (1 - \theta\delta t)^{N_\alpha - N_\alpha^f} \tag{A.2}$$

where $N_\alpha$ is the number of variable nodes in the cluster, i.e., $N_\alpha = 1$ if $\alpha \in \mathcal{R}_L$ else $N_\alpha = 2$ if $\alpha \in \mathcal{R}_S$; $N_\alpha^f$ is the number of variable nodes that have flipped between times

$t$ and $t + \delta t$; and $\mathcal{H}_\alpha(\vec{x}_\alpha)$ is the *cluster hamiltonian* given by:

$$\mathcal{H}_\alpha(\vec{x}_\alpha) = \Big\{ \begin{array}{ll} -\mu_0 H \sum_{i=\{\pm 1\}} i^{\mathbf{1}_{\{\vec{x}_\alpha = i\}}} & \forall \vec{x}_\alpha \in \mathcal{R}_S \\ \sum_{i,j=\{\pm 1\}} \{-\mu_0 H(i+j) - Jij\}^{\mathbf{1}_{\{\vec{x}_\alpha = \{i,j\}\}}} & \forall \vec{x}_\alpha \in \mathcal{R}_L \end{array} \tag{A.3}$$

The key simplification to obtain the original Kikuchi PPF from the general formulation is to assume that the clusters are indistinguishable. This implies that all clusters of same dimensions have the same initial probability distribution as well as the same path variables clusters as well as the joint probability distributions. In other terms,

$$b_\alpha(\mathbf{x}^t{}_\alpha = i) = x_i \quad \forall \alpha \in \mathcal{R}_S; \tag{A.4}$$

$$b_\alpha(\mathbf{x}^t{}_\alpha = \{i,j\}) = y_{ij} \quad \forall \alpha \in \mathcal{R}_L \tag{A.5}$$

and,

$$b_\alpha(\mathbf{x}^{t+\delta t}{}_\alpha = j, \mathbf{x}^t{}_\alpha = i) = X_{i,j} \quad \forall \alpha \in \mathcal{R}_S \tag{A.6}$$

$$b_\alpha(\mathbf{x}^{t+\delta t}{}_\alpha = \{k,l\}, \mathbf{x}^t{}_\alpha = \{i,j\}) = Y_{ij,kl} \quad \forall \alpha \in \mathcal{R}_L \tag{A.7}$$

where $\mathcal{X}_s = \{x_i, y_{ij}\}$ and $\mathcal{X}_p = \{X_{i,j}, Y_{ij,kl}\}$ are the state and path variables in the Kikuchi formulation respectively and $i, j, k, l \in \{\pm 1\}$. The state entropy $\hat{S}^o(\{b\})$ is given by:

$$\hat{S}^o(\{b\}) = -\big[ \sum_{\alpha \in \mathcal{R}_L} \sum_{\mathbf{x}^{t,\delta t}{}_\alpha} b_\alpha(\mathbf{x}^{t,\delta t}{}_\alpha) \ln b_\alpha(\mathbf{x}^t{}_\alpha) + \sum_{\alpha \in \mathcal{R}_S} (1-z) \sum_{\mathbf{x}^{t,\delta t}{}_\alpha} b_\alpha(\mathbf{x}^{t,\delta t}{}_\alpha) \ln b_\alpha(\mathbf{x}^t{}_\alpha) \big] \tag{A.8}$$

$$= -\big[ \sum_{\alpha \in \mathcal{R}_L} \sum_{\mathbf{x}^t{}_\alpha} b_\alpha(\mathbf{x}^t{}_\alpha) \ln b_\alpha(\mathbf{x}^t{}_\alpha) + \sum_{\alpha \in \mathcal{R}_S} (1-z) \sum_{\mathbf{x}^t{}_\alpha} b_\alpha(\mathbf{x}^t{}_\alpha) \ln b_\alpha(\mathbf{x}^t{}_\alpha) \big] \tag{A.9}$$

$$= -\big[ \frac{Nz}{2} \sum_{i,j \in \{\pm 1\}} y_{ij} \ln y_{ij} + N(1-z) \sum_{i \in \{\pm 1\}} x_i \ln x_i \big] \tag{A.10}$$

In the above derivation, we have implicitly used (3.26), which reduces $S^o(\{b\})$ to a constant. The path entropy $\hat{S}^p(\{b\})$ is given by:

$$
\begin{aligned}
\hat{S}^p(\{b\}) &= -[\sum_{\alpha \in \mathcal{R}_L} \sum_{\mathbf{x}^{t,\delta t}{}_\alpha} b_\alpha(\mathbf{x}^{t,\delta t}{}_\alpha) \ln b_\alpha(\mathbf{x}^{t,\delta t}{}_\alpha) + \sum_{\alpha \in \mathcal{R}_S} (1-z) \sum_{\mathbf{x}^{t,\delta t}{}_\alpha} b_\alpha(\mathbf{x}^{t,\delta t}{}_\alpha) \ln b_\alpha(\mathbf{x}^{t,\delta t}{}_\alpha)] \quad \text{(A.11)} \\
&= -[\frac{Nz}{2} \sum_{i,j,k,l \in \{\pm 1\}} Y_{ij,kl} \ln Y_{ij,kl} + N(1-z) \sum_{i \in \{\pm 1\}} X_{i,j} \ln X_{i,j}] \quad \text{(A.12)}
\end{aligned}
$$

The region graph formulation and the associated counting numbers have the single counting property such that each factor and variable is counted exactly once. This property and the indistinguishable property result in the path energy as :

$$
\begin{aligned}
H^p(\{b\}) &= \sum_{\alpha \in \mathcal{R}} c_\alpha \sum_{\mathbf{x}^{t,\delta t}{}_\alpha} b_\alpha(\mathbf{x}^{t,\delta t}{}_\alpha) \ln H_\alpha(\mathbf{x}^{t+\delta t}{}_\alpha | \mathbf{x}^t{}_\alpha) \quad \text{(A.13)} \\
&= -\mathcal{H}(\mathbf{x}^{t+\delta t}) + \mathcal{H}(\mathbf{x}^t) + N \sum_{i=\{\pm 1\}} \{X_{i,-i} \ln(\theta \delta t) + X_{i,i} \ln(1-\theta \delta t)\} \quad \text{(A.14)} \\
&= -\mathcal{H}(\mathbf{x}^{t+\delta t}) + \mathcal{H}(\mathbf{x}^t) + N_f \ln(\theta \delta t) + (N - N_f) \ln(1-\theta \delta t) \quad \text{(A.15)}
\end{aligned}
$$

The term $\{-\mathcal{H}(\mathbf{x}^{t+\delta t}) + \mathcal{H}(\mathbf{x}^t)\}$ denotes the change in energy between current and next state and is expressed in term of path variables as:

$$
\frac{1}{N}\{-\mathcal{H}(\mathbf{x}^{t+\delta t}) + \mathcal{H}(\mathbf{x}^t)\} = \frac{z}{2}J \sum_{i,j,k,l} Y_{ij,kl}(kl - ij) + \mu_0 H \sum_{i,j} X_{i,j}(j - i) \quad \text{(A.16)}
$$

We can write the variational free energy as:

$$
\begin{aligned}
\mathcal{F} &= \hat{S}^p(\{b\}) - \hat{S}^o(\{b\}) + H^p(\{b\}) \quad \text{(A.17)} \\
&= \sum_{i=\pm 1} \{X_{i,-i} \ln(\theta \delta t) + X_{i,i} \ln(1-\theta \delta t)\} - \frac{1}{N}\{\mathcal{H}(\mathbf{x}^{t+\delta t}) + \mathcal{H}(\mathbf{x}^t)\} \\
&\quad - \frac{Nz}{2} \sum_{i,j,k,l \in \{\pm 1\}} Y_{ij,kl} \ln Y_{ij,kl} - N(1-z) \sum_{i \in \{\pm 1\}} X_{i,j} \ln X_{i,j} \\
&\quad + [\frac{Nz}{2} \sum_{i,j \in \{\pm 1\}} y_{ij} \ln y_{ij} + N(1-z) \sum_{i \in \{\pm 1\}} x_i \ln x_i] \quad \text{(A.18)}
\end{aligned}
$$

which is minimized subject to the constraints (3.26) and (3.27). The logarithmic form of the original path probability function $\mathcal{P}$ is given by

$$
\begin{aligned}
\frac{1}{N}\ln\mathcal{P}(\mathcal{X}_p) \;=\;& \sum_{i=\pm 1}\{X_{i,-i}\ln(\theta\delta t) + X_{i,i}\ln(1-\theta\delta t)\} - \frac{1}{N}\{\mathcal{H}(\mathbf{x}^{t+\delta t})+\mathcal{H}(\mathbf{x}^t)\} \\
&-\frac{z}{2}\sum_{i,j,k,l\in\{\pm 1\}}Y_{ij,kl}\ln Y_{ij,kl} - (1-z)\sum_{i\in\{\pm 1\}}X_{i,j}\ln X_{i,j} + \text{const}
\end{aligned}
\tag{A.19}
$$

by using the Stirling approximation and neglecting constant terms. The above term is minimized w.r.t. the path variables, subject to the normalization and consistency constraints.

Comparing (A.18) with (A.19), and observing that $S^o(\{b\})$ reduces to a constant value due to (3.26), we find that the PPF formulation is equivalent to the DynBP formulation using Bethe-Pierls approximation and the trial conditional probability distribution given by:

$$
\hat{p}(\mathbf{x}^{t+\delta t}|\mathbf{x}^t) \propto \exp\{-\mathcal{H}(\mathbf{x}^{t+\delta t})+\mathcal{H}(\mathbf{x}^t)\} \times (\theta\delta t)^{N^f}(1-\theta\delta t)^{N-N^f}
\tag{A.20}
$$

This completes the proof.

# Appendix B

# Derivation for DynBP

In this section, we present the derivation for message update equations of the DynBP algorithm. The DynBP optimization problem is:

$$\min_{b_\alpha(\mathbf{x}^{t,\delta t}{}_\alpha)} \quad \sum_{\alpha \in \mathcal{R}} c_\alpha \{ \sum_{\mathbf{x}^{t,\delta t}{}_\alpha} b_\alpha(\mathbf{x}^{t,\delta t}{}_\alpha) \{ \ln b_\alpha(\mathbf{x}^{t,\delta t}{}_\alpha) - \sum_{a \in \alpha} \ln f_a(\mathbf{x}^{t+\delta t}{}_\alpha | \mathbf{x}^t{}_\alpha) - \ln b_\alpha(\mathbf{x}^t{}_\alpha) \} \} \tag{B.1}$$

$$s.t. \qquad \sum_{\mathbf{x}^{t+\delta t}{}_\alpha} b_\alpha(\mathbf{x}^{t,\delta t}{}_\alpha) = b_\alpha(\mathbf{x}^t{}_\alpha) \quad \forall \alpha \tag{B.2}$$

$$\sum_{\mathbf{x}^{t,\delta t}{}_{\alpha \backslash \beta}} b_\alpha(\mathbf{x}^{t,\delta t}{}_\alpha) = b_\beta(\mathbf{x}^{t,\delta t}{}_\beta) \quad \forall \beta \subseteq \alpha \tag{B.3}$$

Writing the lagrangian, we get

$$
\begin{aligned}
\mathcal{L} \;=\; & \sum_{\alpha \in \mathcal{R}} c_\alpha \sum_{\mathbf{x}^{t,\delta t}{}_\alpha} b_\alpha(\mathbf{x}^{t,\delta t}{}_\alpha) \{ \ln b_\alpha(\mathbf{x}^{t,\delta t}{}_\alpha) + H_\alpha(\mathbf{x}^{t,\delta t}{}_\alpha) - \ln b_\alpha(\mathbf{x}^t{}_\alpha) \} \\
& + \sum_{\alpha \in \mathcal{R}} \sum_{\mathbf{x}^t{}_\alpha} \lambda_\alpha(\mathbf{x}^t{}_\alpha) \{ b_\alpha(\mathbf{x}^t{}_\alpha) - \sum_{\mathbf{x}^{t+\delta t}{}_\alpha} b_\alpha(\mathbf{x}^{t,\delta t}{}_\alpha) \} \\
& + \sum_{\beta \subset \alpha \in \mathcal{R}} \sum_{\mathbf{x}^{t,\delta t}{}_\beta} \lambda_{\beta \rightarrow \alpha}(\mathbf{x}^{t,\delta t}{}_\beta) \{ b_\beta(\mathbf{x}^{t,\delta t}{}_\beta) - \sum_{\mathbf{x}^{t,\delta t}{}_{\alpha \backslash \beta}} b_\alpha(\mathbf{x}^{t,\delta t}{}_\alpha) \}
\end{aligned} \tag{B.4}
$$

Differentiating $\mathcal{L}$ w.r.t $b_\alpha(\mathbf{x}^{t,\delta t}{}_\alpha)$ and setting it to zero, we get:

$$c_\alpha \{ \ln b_\alpha(\mathbf{x}^{t,\delta t}{}_\alpha) + 1 + H_\alpha(\mathbf{x}^{t+\delta t}{}_\alpha | \mathbf{x}^t{}_\alpha) - \ln b_\alpha(\mathbf{x}^t{}_\alpha) \}$$

$$-\lambda_\alpha(\mathbf{x}^t{}_\alpha) + \sum_{\alpha \subset \gamma} \lambda_{\alpha \rightarrow \gamma}(\mathbf{x}^{t,\delta t}{}_\alpha) - \sum_{\beta \subset \alpha} \lambda_{\beta \rightarrow \alpha}(\mathbf{x}^{t,\delta t}{}_\beta) \;=\; 0$$

which results in the update equation for $b_\alpha(\mathbf{x}^{t,\delta t}{}_\alpha)$ of the form,

$$
\begin{aligned}
\ln b_\alpha(\mathbf{x}^{t,\delta t}{}_\alpha) &= -1 - H_\alpha(\mathbf{x}^{t+\delta t}{}_\alpha | \mathbf{x}^t{}_\alpha) + \ln b_\alpha(\mathbf{x}^t{}_\alpha) + \frac{1}{c_\alpha} \lambda_\alpha(\mathbf{x}^t{}_\alpha) \\
&\quad + \frac{1}{c_\alpha} \{ \sum_{\beta \subset \alpha} \lambda_{\beta \to \alpha}(\mathbf{x}^{t,\delta t}{}_\beta) - \sum_{\alpha \subset \gamma} \lambda_{\alpha \to \gamma}(\mathbf{x}^{t,\delta t}{}_\alpha) \}
\end{aligned}
\tag{B.5}
$$

or the equivalent,

$$
b_\alpha(\mathbf{x}^{t,\delta t}{}_\alpha) = \frac{b_\alpha(\mathbf{x}^t{}_\alpha) \prod_{a \in \alpha} f_a(\mathbf{x}^{t+\delta t}{}_a | \mathbf{x}^t{}_a)}{m_\alpha(\mathbf{x}^t{}_\alpha)^{-1/c_\alpha} \times e} \left\{ \frac{\prod_{\beta \subset \alpha} m_{\beta \to \alpha}(\mathbf{x}^{t,\delta t}{}_\beta)}{\prod_{\alpha \subset \gamma} m_{\alpha \to \gamma}(\mathbf{x}^{t,\delta t}{}_\alpha)} \right\}^{1/c_\alpha}
\tag{B.6}
$$

where $m_{\beta \to \alpha}(\mathbf{x}^{t,\delta t}{}_\beta)$ and $m_{\alpha \to \gamma}(\mathbf{x}^{t,\delta t}{}_\alpha)$ are the messages corresponding to the child region $\beta$ and parent region $\gamma$ of region $\alpha$ respectively, given by:

$$
m_{\alpha \to \gamma}(\mathbf{x}^{t,\delta t}{}_\alpha) = \exp\{\lambda_{\beta \to \alpha}(\mathbf{x}^{t,\delta t}{}_\beta)\}
\tag{B.7}
$$

$$
m_{\alpha \to \gamma}(\mathbf{x}^{t,\delta t}{}_\alpha) = \exp\{\lambda_{\alpha \to \gamma}(\mathbf{x}^{t,\delta t}{}_\alpha)\}
\tag{B.8}
$$

for each joint state $\mathbf{x}^{t,\delta t}{}_\alpha$. The message $m_\alpha(\mathbf{x}^t{}_\alpha)$ corresponding to the past state $\mathbf{x}^t{}_\alpha$ is given by:

$$
m_\alpha(\mathbf{x}^t{}_\alpha) = \exp\{\lambda_\alpha(\mathbf{x}^t{}_\alpha)\}
\tag{B.9}
$$

In order to get the iterative update equation for $m_\alpha(\mathbf{x}^t{}_\alpha)$, we observe from (B.6) that at the $i$-th iteration,

$$
\sum_{\mathbf{x}^{t+\delta t}{}_\alpha} b_\alpha^{(i)}(\mathbf{x}^{t,\delta t}{}_\alpha) \propto m_\alpha^{(i)}(\mathbf{x}^t{}_\alpha)^{1/c_\alpha}
\tag{B.10}
$$

Comparing (B.10) with (3.26), we get:

$$
\frac{\sum_{\mathbf{x}^{t+\delta t}{}_\alpha} b_\alpha^{(i)}(\mathbf{x}^{t,\delta t}{}_\alpha)}{b_\alpha(\mathbf{x}^t{}_\alpha)} = \frac{m_\alpha^{(i)}(\mathbf{x}^t{}_\alpha)^{1/c_\alpha}}{m_\alpha^{(i+1)}(\mathbf{x}^t{}_\alpha)^{1/c_\alpha}}
\tag{B.11}
$$

$$
\Rightarrow m_\alpha^{(i+1)}(\mathbf{x}^t{}_\alpha) = m_\alpha^{(i)}(\mathbf{x}^t{}_\alpha) \left\{ \frac{b_\alpha(\mathbf{x}^t{}_\alpha)}{\sum_{\mathbf{x}^{t+\delta t}{}_\alpha} b_\alpha^{(i)}(\mathbf{x}^{t,\delta t}{}_\alpha)} \right\}^{c_\alpha}
\tag{B.12}
$$

Similarly for $m_{\beta \to \alpha}(\mathbf{x}^{t,\delta t}{}_\beta)$, we observe from (B.6) that at the $i$-th iteration,

$$\sum_{\mathbf{x}^{t,\delta t}{}_{\alpha \backslash \beta}} b_\alpha^{(i)}(\mathbf{x}^{t,\delta t}{}_\alpha) \quad \propto \quad m_{\beta \to \alpha}^{(i)}(\mathbf{x}^{t,\delta t}{}_\beta)^{1/c_\alpha} \tag{B.13}$$

$$b_\beta^{(i)}(\mathbf{x}^{t,\delta t}{}_\beta) \quad \propto \quad m_{\beta \to \alpha}^{(i)}(\mathbf{x}^{t,\delta t}{}_\beta)^{-1/c_\beta} \tag{B.14}$$

$$\Rightarrow \frac{\sum_{\mathbf{x}^{t,\delta t}{}_{\alpha \backslash \beta}} b_\alpha^{(i)}(\mathbf{x}^{t,\delta t}{}_\alpha)}{b_\beta^{(i)}(\mathbf{x}^{t,\delta t}{}_\beta)} \quad \propto \quad m_{\beta \to \alpha}^{(i)}(\mathbf{x}^{t,\delta t}{}_\beta)^{1/c_\alpha + 1/c_\beta} \tag{B.15}$$

Comparing (B.15) with (3.27), we get

$$\frac{\sum_{\mathbf{x}^{t,\delta t}{}_{\alpha \backslash \beta}} b_\alpha^{(i)}(\mathbf{x}^{t,\delta t}{}_\alpha)/b_\beta^{(i)}(\mathbf{x}^{t,\delta t}{}_\beta)}{1.0} \quad = \quad \frac{m_{\beta \to \alpha}^{(i)}(\mathbf{x}^{t,\delta t}{}_\beta)^{1/c_\alpha + 1/c_\beta}}{m_{\beta \to \alpha}^{(i+1)}(\mathbf{x}^{t,\delta t}{}_\beta)^{1/c_\alpha + 1/c_\beta}}$$

$$\Rightarrow m_{\beta \to \alpha}^{(i+1)}(\mathbf{x}^{t,\delta t}{}_\beta) \quad = \quad m_{\beta \to \alpha}^{(i)}(\mathbf{x}^{t,\delta t}{}_\beta) \Big\{ \frac{b_\beta^{(i)}(\mathbf{x}^{t,\delta t}{}_\beta)}{\sum_{\mathbf{x}^{t,\delta t}{}_{\alpha \backslash \beta}} b_\alpha^{(i)}(\mathbf{x}^{t,\delta t}{}_\alpha)} \Big\}^{\frac{c_\alpha c_\beta}{c_\alpha + c_\beta}} \tag{B.16}$$

Thus, we get the updated messages using (3.29) and (3.30), which are then used in (B.6) to get the next iteration values for $b_\alpha^{(i+1)}(\mathbf{x}^{t,\delta t}{}_\alpha)$. This completes the proof for DynBP.